\documentclass[runningheads]{llncs}

\usepackage{accv}

\usepackage{accvabbrv}

\usepackage{graphicx}
\usepackage{booktabs}

\usepackage[accsupp]{axessibility}  

\usepackage{hyperref}

\usepackage{graphicx}
\usepackage{booktabs}
\usepackage{array}      
\usepackage{orcidlink}
\usepackage{algorithm}
\usepackage{algpseudocode}

\begin{document}

\title{Masked Feature Encoding for Large-Scale Whole Slide Image Representation} 

\titlerunning{MFE-MIL}

\author{Haoyu He\inst{1} \and
Basile Tessier-Cloutier\inst{3} \and
Yang Wang\inst{1,2} \and
Mahdi S.~Hosseini\inst{1,2,3}}

\authorrunning{H.~He et al.}

\institute{Department of Computer Science and Software Engineering (CSSE),
Concordia University, Montreal, Canada\\
\email{h\_haoy@live.concordia.ca}, \email{yang.wang@concordia.ca},
\email{mahdi.hosseini@concordia.ca} \and
Mila -- Quebec AI Institute, Montreal, Canada \and
McGill University, Montreal, Canada\\
\email{basile.tessiercloutier@mcgill.ca}}

\maketitle

\begin{abstract}

Whole slide image (WSI) analysis in computational pathology follows a multiple instance learning (MIL) pipeline where patch embeddings are extracted independently and aggregated for slide-level prediction, but within-slide variance from staining, scanner, and local texture can overwhelm the discriminative signal.
We propose \textbf{Masked Feature Encoding for Multiple Instance Learning (MFE-MIL)}, a feature-space masking framework that trains a lightweight MLP adapter jointly with a window-based masked reconstruction branch and a MIL classification head.
The two objectives are complementary. Classification guides the adapter to suppress within-slide patch variance, while window-based masked reconstruction provides an auxiliary regularizer for the adapted features without using patch coordinates, coordinate graphs, or segmentation preprocessing. The raster patch-extraction order is used only as a weak implicit prior.
At inference, the decoder is removed, leaving only the adapter and MIL head.
Across CAMELYON16/17, PANDA, and TCGA-BRCA with four diverse encoders, MFE-MIL improves ACC/F1 for nearly all tested aggregator--encoder settings and AUC in most, outperforms coordinate-based spatial methods (CAMIL), and achieves higher AUC than 2DMamba on three of four datasets (UNI). On five TCGA survival cohorts it improves the average concordance index for every aggregator tested, its most consistent gain.
Code is available at \url{https://github.com/AtlasAnalyticsLab/MFE-MIL}.

\end{abstract}

\section{Introduction}
\label{sec:intro}
\begin{figure}[htp]
  \centering
  \small
    \includegraphics[width=1\linewidth]{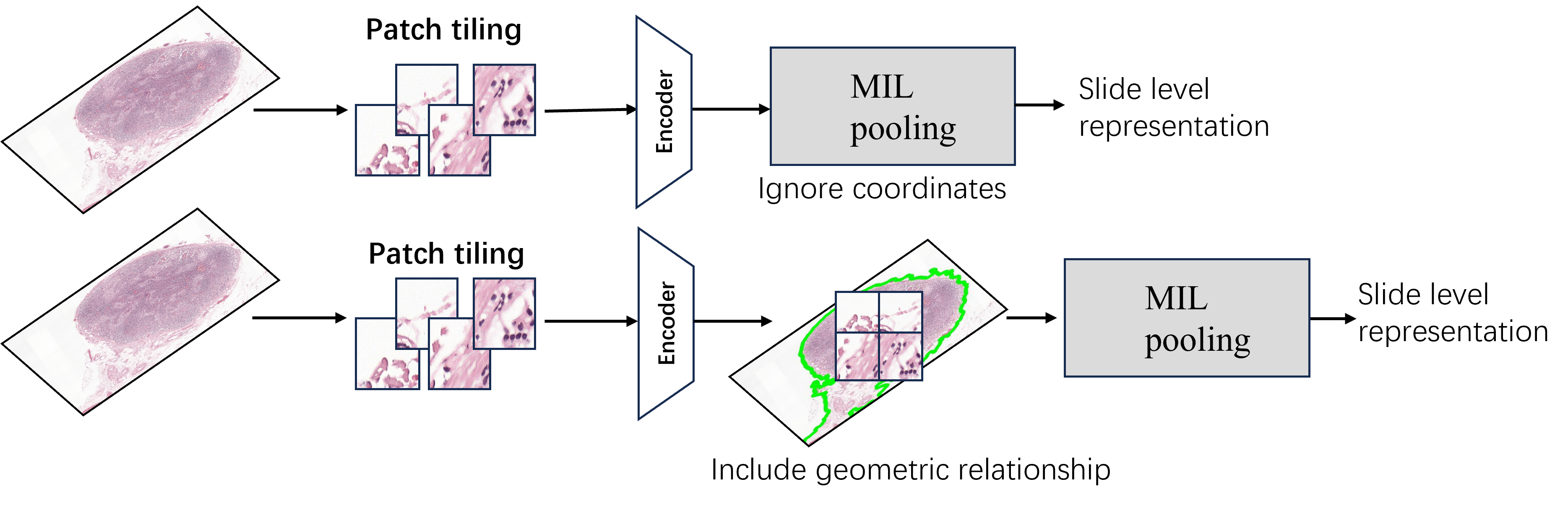}
    \caption{Top: Standard MIL aggregates independently embedded patches whose features carry high within-slide variance. Bottom: MFE-MIL adapts the patch features and regularizes them with masked reconstruction over neighbouring packed features, giving a more discriminative slide-level representation.}
    \label{fig:intro}
\end{figure}

In computational pathology (CPath)~\cite{cifci2023ai,hosseini2024computational, song2023artificial}, Whole Slide Image (WSI) classification is typically achieved through Multiple Instance Learning (MIL)~\cite{wang2019weakly,lu2021data}, where a frozen encoder extracts patch-level embeddings, and an aggregator combines them into a slide-level prediction.
Most MIL methods treat patches as an unordered set, so each embedding is computed in isolation and carries within-slide variation (staining, scanner, local texture) that can mask the diagnostic signal (\cref{fig:intro}).

\begin{figure*}[t]
  \centering
  \includegraphics[width=\linewidth]{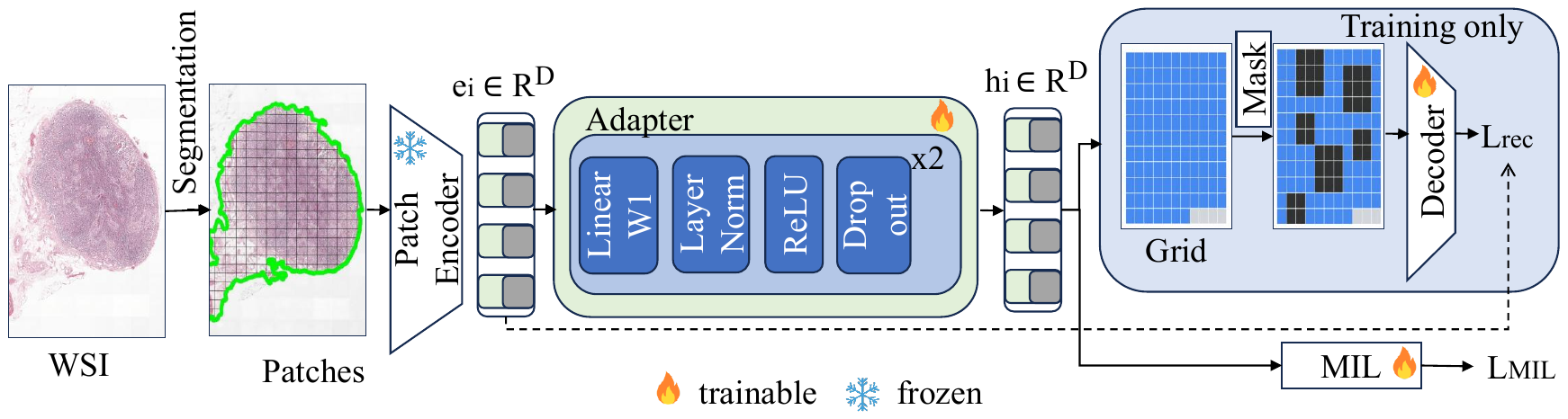}
  \caption{Overview of MFE-MIL. Patch features $e_i$ from a frozen encoder are mapped by a trainable two-block MLP adapter (Linear--LayerNorm--ReLU--Dropout, $\times 2$) to $h_i$. The adapted features feed a MIL head, trained with $\mathcal{L}_\text{MIL}$, and, during training only, a reconstruction branch: features are packed in extraction order into a grid, contiguous windows are masked, and a Transformer decoder reconstructs the $\ell_2$-normalized frozen embeddings $\bar{e}_i$ of the masked patches ($\mathcal{L}_\text{rec}$; dashed arrow: reconstruction target). No patch coordinates are used, and the decoder is discarded at inference.}
  \label{fig:arch}
\end{figure*}

Recent spatially aware MIL methods try to close this gap by incorporating explicit structural priors such as coordinate grids, topology graphs, or segmentation
masks~\cite{fourkioti2024camil,woelflein2023dasmil,zhang2025_2dmamba,fang2024sammil},
but require tight encoder-aggregator coupling that reduces compatibility with standard frozen-feature pipelines.
We propose \textbf{MFE-MIL}, built on the hypothesis that high within-slide variance in independently computed patch embeddings, rather than missing spatial priors, limits MIL aggregators.
An architecture-agnostic MLP adapter, jointly trained with MIL classification and a window-based masked reconstruction objective, learns
slide-coherent representations: the MIL classification objective drives the adapter to
suppress within-slide patch variance, 
while the window MFE pretext task acts as an auxiliary reconstruction regularizer on the compressed space, giving all aggregators a cleaner discriminative signal
(\cref{fig:arch}).
Unlike MAE-based CPath approaches, the reconstruction acts at fine-tuning time rather than during backbone pretraining (\cref{sec:related}).

\noindent\textbf{Contributions:}
\begin{itemize}
    \item We introduce MFE-MIL, a plug-and-play masked feature reconstruction framework for frozen-encoder MIL pipelines that requires no patch coordinates at either training or inference: it packs features in tissue-extraction order and applies windowed masked reconstruction as a lightweight regularizer, treating raster order only as a weak implicit prior.

    \item We characterize the underlying mechanism. Classification suppresses within-slide patch variance, while window-based masked reconstruction further regularizes the adapted features. Controlled experiments using shuffled order, true coordinates, 1D masking, distance-matched masking, and single-token masks show that improvements in feature coherence do not consistently depend on the spatial arrangement of the mask, although standard window masking achieves the highest AUC on CAMELYON16.
    
    \item Across CAMELYON16/17, PANDA, and TCGA-BRCA with four encoders and six aggregators, MFE-MIL improves ACC/F1 in nearly all settings, outperforms CAMIL without coordinate graphs, and measurably sharpens tumour localization (ABMIL attention-localization AUROC $0.879\!\to\!0.942$ against CAMELYON16 annotations); gains are largest for the weakest-baseline aggregators, consistent with the variance-suppression mechanism.
\end{itemize}




\section{Related Work}
\label{sec:related}

\noindent\textbf{MIL in CPath.}
MIL treats a WSI as a bag of patch embeddings with only slide-level
supervision~\cite{wang2019weakly,lu2021data}. Aggregation methods range from
simple pooling to attention-based aggregation (ABMIL~\cite{ilse2018attention},
CLAM~\cite{lu2021data}) and interaction-aware variants such as
DSMIL~\cite{li2021dsmil}, TransMIL~\cite{shao2021transmil}, and
DTFD-MIL~\cite{zhang2022dtfd}. However, most methods operate on patches primarily as an unordered set, leaving spatial relationships between tissue regions largely unexploited. This is particularly limiting for WSI analysis, where diagnostically relevant patterns can depend on the spatial organization and local context of tissue regions.

\noindent\textbf{Spatially aware MIL.}
Recent methods incorporate explicit spatial priors via coordinate-aware or topology-aware modeling~\cite{zhu2022setmil,fourkioti2024camil,
woelflein2023dasmil,zhang2025_2dmamba} or segmentation
priors~\cite{fang2024sammil}. These methods often require additional preprocessing (\eg, coordinates, graphs, segmentation~\cite{kirillov2023sam})
or specialized architectural coupling, reducing plug-and-play compatibility with existing frozen-feature pipelines; graph-based variants (e.g., WiKG~\cite{li2024wikg}) and recent expert-clustering methods~\cite{expertcluster2026} add further context modules but still require graph or clustering construction. MFE-MIL takes a different approach, where a masked-reconstruction objective regularizes the adapted features rather than baking spatial priors into the architecture, so the aggregator interface remains unchanged.

\noindent\textbf{Foundation models for pathology.}
Pathology-pretrained encoders such as CONCH~\cite{lu2024conch}, UNI~\cite{chen2024uni}, ViT-S/16-SSL~\cite{kang2023}, GigaPath~\cite{xu2024gigapath}, PRISM~\cite{shaikovski2024prism}, and TITAN~\cite{ding2025titan} have become a common starting point for WSI analysis. In standard pipelines, these encoders are kept frozen and task-specific MIL aggregators are trained on the resulting patch embeddings. This setting avoids the substantial cost of retraining large foundation models while allowing the downstream model to adapt the representations to the target task. MFE-MIL is fully compatible with this frozen-encoder setup and requires no backbone retraining.

\noindent\textbf{Masked autoencoders and auxiliary tasks.}
MAE-style pretraining~\cite{he2022mae}, self-supervised feature learning~\cite{mammadov2025selfsupervision}, and masked-reconstruction pre-training~\cite{lu2023multimodal,wu2024pancancer} are widely used in CPath, but typically shape a generic representation during backbone pretraining rather than a task-aligned one.
MFE-MIL applies masked feature
reconstruction jointly with MIL classification at fine-tuning time. The most closely related prior work is MHIM-MIL~\cite{tang2023mhim}, which introduces masking during MIL training but selects patches by attention score
(hard instance mining), whereas MFE-MIL masks contiguous blocks. As we show in \cref{sec:analysis}, this masked-reconstruction objective acts as a regularizer whose effect on feature coherence does not reliably depend on the spatial arrangement or grouping of the mask; the coherence gap is comparable across shuffled-order, true-coordinate, 1D, distance-matched, and single-token controls.

\section{Methodology}
\label{sec:method}

High within-slide patch variance (staining, scanner, local texture) can
overwhelm the discriminative signal, and frozen encoders further encode physical
artifacts (folds, pen marks, bubbles) as distinctive, drawing attention to
non-pathological regions.
Rather than encoding space architecturally as prior methods
do~\cite{fourkioti2024camil}, MFE-MIL shapes the feature space through the
training objective: the MIL classification loss compresses within-slide variance,
and a jointly optimized window-based masked-reconstruction pretext task provides an auxiliary
reconstruction regularizer on the adapted features.
The attention entropy in \cref{tab:attn_entropy} shows that this compression
sharpens tumour attention rather than flattening it.

\noindent\textbf{Overall architecture.}
Given a WSI bag $\mathcal{X}=\{x_i\}_{i=1}^N$, a frozen pretrained encoder
$f_\text{enc}$ produces patch embeddings $e_i = f_\text{enc}(x_i)$;
a two-layer MLP adapter
$g_\theta$ maps these into a task-specific latent space feeding two branches:
\begin{equation}
  \underbrace{h_i = g_\theta(e_i)}_{\text{adapter}},\quad
  \underbrace{\hat{\tilde{H}} = f_\text{dec}(\tilde{H}', M, P)}_{\text{reconstruction (train only)}},\quad
  \underbrace{\hat{Y} = f_\text{class}(f_\text{agg}(H))}_{\text{MIL head}}.
  \label{eq:overview}
\end{equation}
The decoder is discarded at inference, leaving only the adapter and MIL head.

\noindent\textbf{Task-aware MLP adapter.}
In the spirit of parameter-efficient fine-tuning~\cite{han2024peft,zhang2025peft}, the adapter is a two-block MLP, $g_\theta=\phi_2\circ\phi_1$, where each block applies a linear layer, layer normalization, ReLU, and dropout:
\begin{equation}
  \phi_k(x) = \text{Dropout}_{0.25}\!\bigl(\text{ReLU}\bigl(\text{LN}(W_k x + b_k)\bigr)\bigr),
  \qquad W_k\!\in\!\mathbb{R}^{D\times D},
  \label{eq:adapter}
\end{equation}
where $D$ is the frozen encoder output dimension ($384$, $512$, $1024$, and $1024$ for ViT-S/16-SSL, CONCH, UNI, and ViT-L/16-IN21K). The adapter preserves the feature dimension ($d\!=\!D$), so it is a drop-in replacement with no architectural change downstream; for UNI it adds $2.1$M parameters.

The decoder reconstructs the $\ell_2$-normalized frozen embeddings
$\bar{e}_i=e_i/\|e_i\|_2$ of the masked patches from the adapted features of the
visible ones.
The target is a fixed function of the frozen encoder and does not change during
training, so the reconstruction loss cannot be reduced by collapsing the adapter
output: if all $h_i$ were a constant, the decoder could not recover the varying
targets and $\mathcal{L}_\text{rec}$ would stay high.
The adapter must therefore keep enough information to predict masked features,
while the classification loss shapes the same representation toward
class-discriminative directions.
The reconstruction branch takes the normalized embeddings $\bar{e}_i$ as adapter
input, whereas the MIL branch takes the unnormalized $e_i$.

\noindent\textbf{Packed grid and masked reconstruction.}
Given bag $\mathcal{X}=\{x_i\}_{i=1}^N$, the frozen encoder produces
embeddings $e_i$ and the adapter produces $H=[h_1,\ldots,h_N]^\top$.
The reconstruction grid is built \emph{without} patch coordinates: the $N$
adapted features are packed in tissue-extraction order into a square grid of
side $R\!=\!C\!=\!\lceil\sqrt{N}\rceil$,
\begin{equation}
  \tilde{H}\bigl[\lfloor k/C\rfloor,\; k \bmod C,\; :\bigr] = h_{k},
  \qquad k=0,\ldots,N-1,
  \label{eq:gridmap}
\end{equation}
giving $\tilde{H}\!\in\!\mathbb{R}^{R\times C\times d}$ (the reconstruction branch packs $\bar{h}_k=g_\theta(\bar{e}_k)$ in place of $h_k$); the trailing
$R\!\cdot\!C\!-\!N$ cells are zero-padded.
Patches are extracted in raster order over the tissue contour, so cells that
are adjacent within a row are often physically adjacent on the slide. This
coordinate-free packing therefore preserves only partial locality: measured
against the true patch coordinates, within-row precision is $0.89$--$0.97$ across
datasets and across-row precision is below $0.05$ (see the supplementary material),
since the correspondence breaks down across rows, at row wraps, and at irregular
tissue boundaries.
The pipeline uses no patch coordinates at either training or
inference; we show in \cref{sec:analysis} that the resulting regularization
does not reliably depend on the spatial arrangement of the mask.

The decoder receives fixed 2D sine-cosine positional embeddings
\begin{equation}
  P_{r,c} = \text{SinCos2D}(r,\,c) \;\in\; \mathbb{R}^{d_\text{dec}}
  \label{eq:posemb}
\end{equation}
for each grid cell, where $d_\text{dec}\!=\!512$ is the decoder width.  Applying structured mask
$M\in\{0,1\}^{R\times C}$ gives
\begin{equation}
  \tilde{H}' = \text{Mask}(\tilde{H}, M), \quad
  \hat{\tilde{H}} = f_\text{dec}(\tilde{H}', M, P).
  \label{eq:recon}
\end{equation}
The reconstruction loss is the mean squared error between the decoder output and
the normalized frozen embeddings $\bar{E}\in\mathbb{R}^{R\times C\times D}$
(packed like $\tilde{H}$) over the masked positions $\mathcal{M}$:
\begin{equation}
  \mathcal{L}_\text{rec} = \frac{1}{|\mathcal{M}|\,D}
  \sum_{(i,j)\in\mathcal{M}}
  \bigl\|\bar{E}_{i,j,:} - \hat{\tilde{H}}_{i,j,:}\bigr\|_2^2,
  \label{eq:lrec}
\end{equation}
where $\mathcal{M}$ denotes the masked cells that hold a real patch;
zero-padded cells may be masked but are excluded from $\mathcal{M}$, so the decoder is
never asked to reconstruct absent tissue.

\noindent\textbf{Window-based adaptive masking.}
To prevent the decoder from trivially interpolating from immediately adjacent patches, we mask blocks that are contiguous in the packed grid rather than individual tokens. We define a square window of radius $w$ centered at grid cell $(p,q)$:
\begin{equation}
  \begin{aligned}
    \mathcal{W}(p,q,w) &= \{(i,j) : |i-p|\leq w,\; |j-q|\leq w\}, \\
    &\quad |\mathcal{W}| = (2w+1)^2,
  \end{aligned}
  \label{eq:window}
\end{equation}
and build $\mathcal{M}$ in two phases.
First, non-overlapping windows of radius $w\in\{1,2\}$ (i.e., $3\!\times\!3$ or
$5\!\times\!5$ blocks, chosen uniformly at random) are placed at random grid
positions until they cover about $80\%$ of the masking budget
$N_\text{target}=\lfloor r\cdot RC\rfloor$, where $RC$ is the number of cells of
the packed grid.
The remaining cells are then masked individually ($w=0$) to reach the budget
exactly.
Masking whole blocks of the packed grid makes the pretext task harder than
masking single cells, and empirically gives the best AUC (\cref{tab:ablation}).

This two-phase strategy (illustrated in the supplementary material)
outperforms any fixed window size (\cref{tab:ablation}).
We set $r\!=\!75\%$ following MAE~\cite{he2022mae}.

\noindent\textbf{Decoder architecture.}
The decoder $f_\text{dec}$ is a lightweight Transformer with a fixed hidden width $d_\text{dec}\!=\!512$ independent of $D$.
A linear layer maps the $D$-dimensional adapted features to $d_\text{dec}$, four Transformer blocks (16 attention heads of dimension 32, feedforward $512\!\to\!2048\!\to\!512$, i.e., ratio 4) process the tokens, and a final LayerNorm and linear layer map back to $D$ dimensions.
This amounts to about $13.7$M parameters for $D\!=\!1024$ ($13.0$M for ViT-S and $13.1$M for CONCH).
Before decoding, each masked grid cell is replaced by a learned \texttt{[MASK]}
token to separate known from unknown positions.
The 2D positional embeddings $P$ are added to all tokens so the decoder can
use absolute grid location during reconstruction.

\noindent\textbf{Joint training objective.}
The MIL head aggregates adapted features for bag-level prediction:
$
  z = f_\text{agg}(H),\quad \hat{Y} = f_\text{class}(z).
$
The total objective balances classification and reconstruction:
$
  \mathcal{L}_\text{total} =
  (1-\lambda_\text{rec})\,\mathcal{L}_\text{MIL}
  + \lambda_\text{rec}\,\mathcal{L}_\text{rec},
  \label{eq:total}
$
where $\lambda_\text{rec}=0.3$ is chosen from $\{0.1, 0.2, 0.3\}$ (supplementary). Only the MLP adapter, decoder, and MIL head are updated; the foundation encoder is frozen.
The MIL branch never uses decoder outputs, so discarding the decoder causes no train--test mismatch; \cref{tab:cam16_class} is computed on adapter outputs alone.
We evaluate $f_\text{agg}$ as: mean/max pooling, ABMIL~\cite{ilse2018attention},
CLAM-SB and CLAM-MB~\cite{lu2021data}, and TransMIL~\cite{shao2021transmil}.
The full training procedure is summarized in the supplementary material.

\noindent\textbf{Plug-and-play aggregator compatibility.}
The adapter maps each patch independently and keeps its dimension, so any aggregator that consumed $\{e_i\}$ can consume $\{h_i\}$ unchanged. At inference the pipeline is
$\text{encoder} \xrightarrow{\text{frozen}} e_i
 \xrightarrow{g_\theta} h_i
 \xrightarrow{f_\text{agg}} z
 \xrightarrow{f_\text{class}} \hat{Y}$,
with no decoder, whereas coordinate-based methods such as CAMIL need explicit spatial inputs.

\section{Experiments}
\label{sec:experiments}

\begin{figure*}[!tbp]
  \centering
  \includegraphics[width=\linewidth]{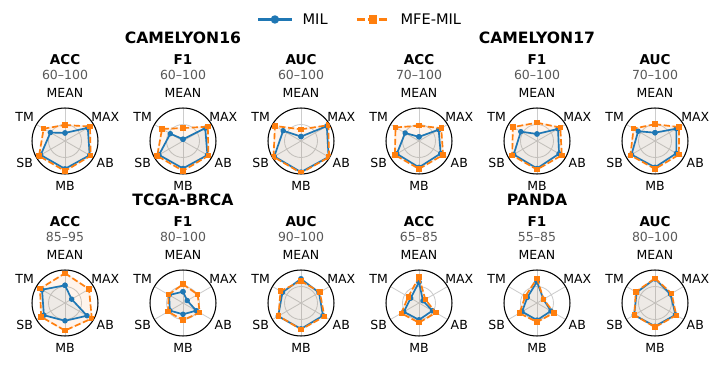}
  \caption{ACC / F1 / AUC (\%) for six MIL aggregators without (solid) and with
  (dashed) MFE-MIL (\textbf{UNI}). Each dataset is a group of three radars; top:
  CAMELYON16, CAMELYON17; bottom: TCGA-BRCA, PANDA. AB = ABMIL, MB = CLAM-MB,
  SB = CLAM-SB, TM = TransMIL. Each radar has its own range (printed above it).
  Means over 3 splits $\times$ 5 seeds; details in the supplementary.}
  \label{fig:radar_uni}
\end{figure*}

\subsection{Experimental Setup}

\noindent\textbf{Datasets.}
CAMELYON16~\cite{bejnordi2017camelyon16} (400 WSIs, tumour detection), CAMELYON17~\cite{bandi2019camelyon17} (500 WSIs, ITC/micro/macro grouped as positive), PANDA~\cite{bulten2022panda} (10,614 prostate biopsies, 6 Gleason grades, 8:1:1), TCGA-BRCA~\cite{tcga} (1,033 slides, IDC vs.\ ILC, 8:1:1).
Five TCGA cohorts~\cite{tcga} (KIRC, KIRP, LUAD, STAD, UCEC) are used for survival.
All WSIs are tiled into 256$\times$256 patches at 20$\times$ with tissue detection and patch extraction performed by AtlasPatch~\cite{alagha2026atlaspatch}. All splits are made at the patient (case) level, so slides from the same patient never span splits.
All experiments use a 3 splits $\times$ 5 seeds setting, and on CAMELYON16 the official 128-slide test set is held fixed across all runs.

\noindent\textbf{Encoders.}
Pathology-pretrained UNI~\cite{chen2024uni}, CONCH~\cite{lu2024conch}, and ViT-S/16-SSL~\cite{kang2023}, plus general-purpose ViT-L/16-IN21K~\cite{dosovitskiy2021image}.

\noindent\textbf{Baselines.}
CAMIL, 2DMamba, and MHIM-MIL were run with their official default hyperparameters on the same split and run protocol. All methods use identical frozen features.

\noindent\textbf{Implementation.}
Classification models are trained with Adam (lr $1\!\times\!10^{-4}$, no weight decay) for up to 200 epochs with early stopping on validation AUC (patience 10; macro one-vs-rest AUC for PANDA), one slide per step (batch size 1), class-balanced sampling of training slides, fp16 mixed precision, masking ratio $r\!=\!75\%$, and $\lambda_\text{rec}\!=\!0.3$. Survival models use Adam (lr $2\!\times\!10^{-4}$, weight decay $1\!\times\!10^{-5}$) with gradient accumulation over 32 slides and up to 20 epochs with early stopping on validation C-index (patience 5). No learning-rate schedule is used, and all runs use a single NVIDIA H100.

\noindent\textbf{Metrics.}
Slide-level ACC / macro-F1 / AUC (one-vs-rest for PANDA's six grades), and C-index for survival.

\subsection{Survival Prediction}
\label{sec:survival}

We first evaluate MFE-MIL on survival prediction, an established weakly-supervised
WSI task typically addressed with attention-MIL or multimodal
histology--genomics models~\cite{chen2021mcat,jaume2024survpath}: the MIL head is
substituted with a discrete-time hazard output over four survival-time bins,
trained with the negative log-likelihood survival loss,
with the adapter and reconstruction branch otherwise unchanged.
\cref{tab:survival} reports concordance index (C-index) on the five TCGA
cohorts (UNI embeddings) for each aggregator with and without MFE-MIL.
Among the spatially aware baselines, only 2DMamba is reported here: CAMIL and
SETMIL provide no survival-prediction implementation, so they are excluded from
this task.

MFE-MIL improves the average C-index (\cref{tab:survival}, AVG column) for every one of the six aggregators
tested, making it the most consistent result in this paper.
Every MFE-MIL variant also exceeds the average C-index of the spatially aware
2DMamba ($0.699$--$0.714$ versus $0.695$), without using patch coordinates.
Paired $t$-tests over all cohort and aggregator pairs confirm the overall
improvement is statistically significant ($p < 0.05$). The largest gains are on UCEC and KIRP, two cohorts with diffuse prognostic signals; STAD gains are small or negative for high-baseline aggregators, while MAX, with the lowest STAD baseline, gains $0.081$. However, MAX's baseline C-index is near-chance on KIRP, LUAD and STAD, so its large gains mainly reflect fixing a weak baseline; the smaller, consistent gains on ABMIL, CLAM-MB, and CLAM-SB are the more informative signal.
On KIRC most aggregators plateau; the exceptions are MAX and CLAM-MB, which rise $0.047$ and $0.011$, and TransMIL, which drops $0.015$, likely because compressing within-slide variance reduces the per-patch diversity that its long-range self-attention relies on.

\begin{table}[!tbp]
\centering
\scriptsize
\caption{Survival prediction C-index on five TCGA cohorts (UNI embeddings).
Results averaged over 3 splits $\times$ 5 seeds (15 runs per cell).
Best result per column in \textbf{bold}; second best \underline{underlined}.}
\label{tab:survival}
\setlength{\tabcolsep}{3.5pt}
\begin{tabular}{@{}ll ccccc r@{}}
\toprule
\multicolumn{2}{@{}l}{Method} & KIRC & KIRP & LUAD & STAD & UCEC & AVG \\
\midrule
MEAN     & MIL     & \underline{0.749} & 0.814 & 0.594 & 0.651 & 0.681 & 0.698 \\
         & +MFE    & 0.748 & 0.835 & \textbf{0.612} & 0.642 & 0.728 & \underline{0.713} \\[2pt]
MAX      & MIL     & 0.708 & 0.552 & 0.538 & 0.532 & 0.644 & 0.595 \\
         & +MFE    & \textbf{0.755} & 0.805 & 0.589 & 0.613 & 0.734 & 0.699 \\[2pt]
ABMIL    & MIL     & 0.740 & 0.821 & 0.601 & 0.639 & 0.608 & 0.682 \\
         & +MFE    & 0.741 & 0.829 & \underline{0.608} & 0.622 & \textbf{0.742} & 0.708 \\[2pt]
CLAM-MB  & MIL     & 0.736 & 0.814 & 0.594 & \underline{0.659} & 0.680 & 0.697 \\
         & +MFE    & 0.747 & 0.840 & 0.598 & \textbf{0.661} & 0.725 & \textbf{0.714} \\[2pt]
CLAM-SB  & MIL     & 0.742 & 0.820 & 0.595 & 0.644 & 0.659 & 0.692 \\
         & +MFE    & 0.739 & \underline{0.844} & 0.598 & 0.636 & 0.726 & 0.708 \\[2pt]
TransMIL & MIL     & 0.732 & 0.807 & 0.573 & 0.623 & 0.706 & 0.688 \\
         & +MFE    & 0.717 & \textbf{0.865} & 0.581 & 0.611 & \underline{0.740} & 0.703 \\
\midrule
\multicolumn{2}{@{}l}{2DMamba} & 0.741 & 0.800 & 0.596 & 0.654 & 0.683 & 0.695 \\
\bottomrule
\end{tabular}
\end{table}

\subsection{Classification Results}
\begin{figure*}[!t]
  \centering
  \includegraphics[width=\linewidth]{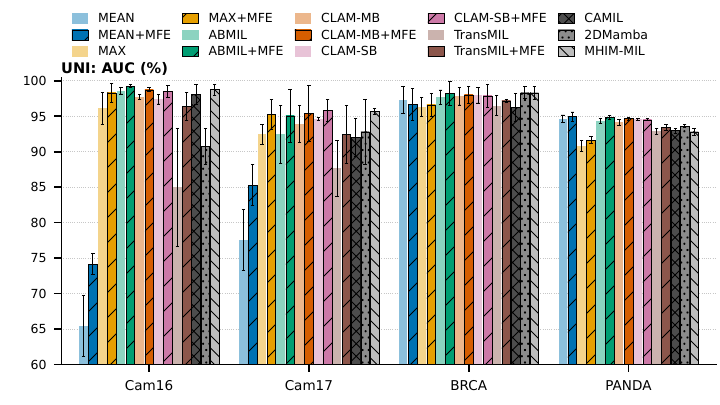}
  \caption{AUC (\%) for all six MIL aggregators (light = base, dark hatched =
  +MFE-MIL), the spatial baselines CAMIL and 2DMamba, and the masking-based MHIM-MIL (gray), using
  \textbf{UNI} embeddings. Error bars: standard deviation over 3 splits
  $\times$ 5 seeds; Y-axis truncated. ACC/F1 and the other encoders are in the
  supplementary.}
  \label{fig:bars_uni}
\end{figure*}

\cref{fig:radar_uni} reports per-aggregator ACC/F1/AUC on all four datasets with UNI embeddings, and \cref{fig:bars_uni} shows the AUC of all six aggregators next to the spatially aware baselines CAMIL~\cite{fourkioti2024camil} and 2DMamba~\cite{zhang2025_2dmamba} and the masking-based MHIM-MIL~\cite{tang2023mhim}. The CONCH counterpart and full numerical results for all four encoders are in the supplementary (SETMIL~\cite{zhu2022setmil} is omitted, as it failed to converge on CAMELYON16/17 with these encoders).

MFE-MIL improves ACC and F1 for every aggregator under the UNI and CONCH encoders and in nearly all ViT-L/ViT-S settings, with gains largest where the baseline is weakest; paired $t$-tests over 15 runs are significant on CAMELYON16, CAMELYON17, and PANDA ($p<0.05$) but not on TCGA-BRCA ($p=0.19$). Against spatially aware baselines, MFE-MIL beats CAMIL on all four datasets without coordinate graphs at inference, and by AUC the best MFE variant leads 2DMamba on all but TCGA-BRCA (within one std); MFE-ABMIL beats MHIM-MIL on ACC and F1 across all datasets and both encoders, and on AUC for two of four datasets under each encoder. AUC regressions are rare and small (at most $0.8$ points), mostly on near-ceiling TCGA-BRCA, where UNI baseline AUC is already $96.3$--$98.0\%$ for every aggregator, leaving too little headroom for a gain to reach significance.

On the general-purpose ViT-L/16-IN21K and the smaller pathology-pretrained ViT-S/16-SSL encoders (full results in the supplementary), MFE-MIL's gains are \emph{larger} than on UNI and CONCH (e.g., ViT-L ABMIL CAMELYON17 AUC $78.7\!\to\!82.8$), consistent with weaker features carrying more within-slide variance to suppress. MFE-MIL also exceeds the recent ACMIL~\cite{zhang2024acmil} aggregator on AUC across all four datasets (supplementary).

The supplementary UNI table isolates the reconstruction branch using a classification-only adapter (frozen $\to$ adapter $\to$ +MFE). For ABMIL,
CLAM-MB, and CLAM-SB, the adapter alone is inconsistent relative to frozen
features, while adding reconstruction improves AUC in every case, most notably on CAMELYON16/17 and more modestly on
near-saturated PANDA; for MAX it is slightly lower on CAMELYON16/17 ($98.62\!\to\!98.31$, $95.53\!\to\!95.31$).

\subsection{Computational Cost}

At inference on CAMELYON16 (UNI), with the decoder discarded, MFE-MEAN adds only
2.1M parameters over MEAN ($2.63$M versus $0.53$M) and $0.12$\,GB of peak memory
($0.60$ versus $0.48$\,GB), raising per-slide inference from $0.010$ to $0.049$\,s
(an aggregator-independent MLP overhead). By contrast, CAMIL needs $14.60$\,GB and
$0.353$\,s for graph construction, and 2DMamba $4.60$\,GB and $0.082$\,s. Training adds the reconstruction decoder ($13.7$M parameters), which raises per-epoch time and peak training memory by ${\sim}6\times$ on CAMELYON16 and end-to-end training time by ${\sim}21\times$; this is a one-time cost, as the decoder is discarded at inference.

\subsection{Ablation Studies}

All ablations use PANDA with UNI embeddings. Reported AUC values use three pre-defined splits $\times$ 5 seeds.

\noindent\textbf{Adapter depth and masking strategy.}
Capacity alone does not explain the gain (\cref{tab:ablation}, top block): MIL-only AUC is non-monotonic in depth and never reaches depth-2 MFE-MIL's performance; adding reconstruction improves every depth, and depth-2 MFE-MIL (our full model) is best overall, though CLAM-SB peaks at depth-1, likely because its single attention branch saturates at lower adapter capacity while deeper adapters slightly overfit its instance loss.
Masking granularity also matters for AUC (\cref{tab:ablation}, bottom block): random single-token and $3\!\times\!3$ masking stall at $93.56$ avg.\ AUC, a fixed $5\!\times\!5$ window reaches $93.88$, and adaptive multi-scale window masking is best, preventing the decoder from exploiting any single predictable block size.

\begin{table}[!tbp]
\centering
\scriptsize
\caption{Ablations on PANDA (UNI), AUC (\%). \textit{Top block}: adapter depth
(MIL-only vs.\ MFE, layers 0--2; depth 0 = no adapter) at adaptive-window masking; depth-2 MFE is our
full model. \textit{Bottom block}: masking strategy at depth-2
(Window\,=\,depth-2 MFE). Best per column \emph{within each block} in \textbf{bold}.}
\label{tab:ablation}
\setlength{\tabcolsep}{4pt}
\begin{tabular}{@{}lcccccr@{}}
\toprule
Setting & MEAN & MAX & AB & MB & SB & Avg \\
\midrule
MIL, 0 & 94.68 & 90.85 & 94.34 & 94.13 & 94.57 & 93.71 \\
MIL, 1 & 94.52 & 91.02 & 94.37 & 94.22 & 94.60 & 93.75 \\
MIL, 2 & 94.27 & 91.10 & 94.23 & 94.24 & 94.21 & 93.61 \\
MFE, 0 & 94.82 & 91.39 & 94.52 & 94.19 & 94.70 & 93.92 \\
MFE, 1 & 94.37 & 91.25 & 94.56 & 94.31 & \textbf{94.89} & 93.88 \\
MFE, 2 & \textbf{94.93} & \textbf{91.62} & \textbf{94.85} & \textbf{94.67} & 94.50 & \textbf{94.11} \\
\midrule
Single-token & 94.58 & 90.71 & 93.95 & 94.21 & 94.37 & 93.56 \\
Small ($3\!\times\!3$) & 94.20 & 91.21 & 94.03 & 94.13 & 94.24 & 93.56 \\
Large ($5\!\times\!5$) & 94.39 & \textbf{91.63} & 94.14 & 94.42 & \textbf{94.84} & 93.88 \\
Window (ours) & \textbf{94.93} & 91.62 & \textbf{94.85} & \textbf{94.67} & 94.50 & \textbf{94.11} \\
\bottomrule
\end{tabular}
\end{table}

\noindent\textbf{Reconstruction loss weight.}
$\lambda_\text{rec}=0.3$ achieves the best average AUC on PANDA across all MIL
heads among the values tested ($0.1$, $0.2$, $0.3$; supplementary material).

\noindent\textbf{Simple normalization baseline.}
A per-slide z-score normalization control does not replicate the learned
adapter: it leaves attention entropy indistinguishable from frozen features
($0.982$ versus $0.985$) and makes the coherence ratio numerically unstable
(supplementary).

\subsection{Qualitative Evaluation}

\begin{figure*}[!t]
\begin{minipage}[t]{0.70\linewidth}
\vspace{0pt}\centering
\newcommand{\hcolw}{0.30\linewidth}%
\setlength{\tabcolsep}{1.5pt}
\begin{tabular}{c@{\hspace{3pt}}ccc}
 & \small\textbf{WSI} & \small\textbf{ABMIL} & \small\textbf{MFE-ABMIL} \\
\rotatebox[origin=c]{90}{\small\textbf{CAM16}} &
\raisebox{-0.5\height}{\includegraphics[width=\hcolw]{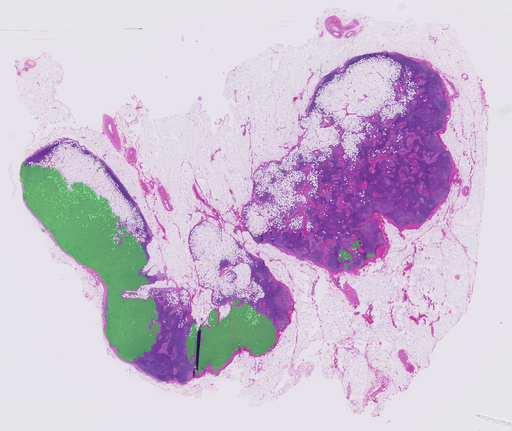}} &
\raisebox{-0.5\height}{\includegraphics[width=\hcolw]{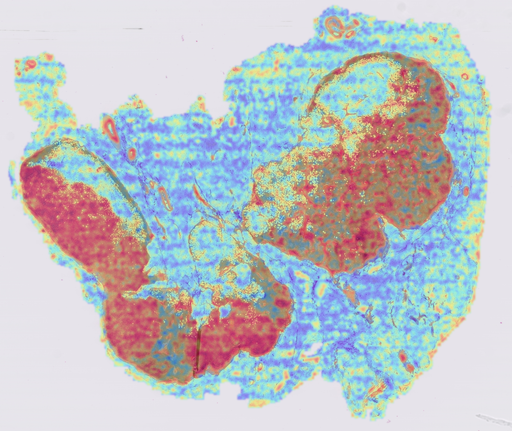}} &
\raisebox{-0.5\height}{\includegraphics[width=\hcolw]{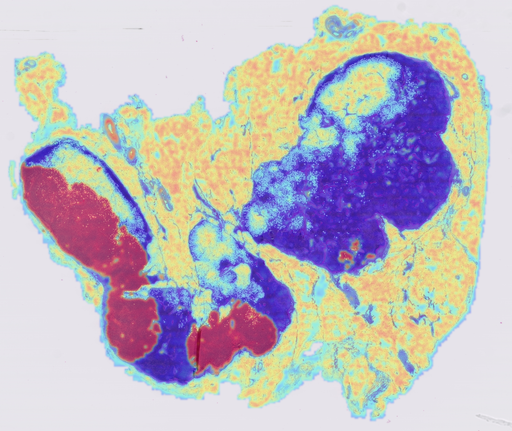}} \\[1pt]
\rotatebox[origin=c]{90}{\small\textbf{LUAD}} &
\raisebox{-0.5\height}{\includegraphics[width=\hcolw]{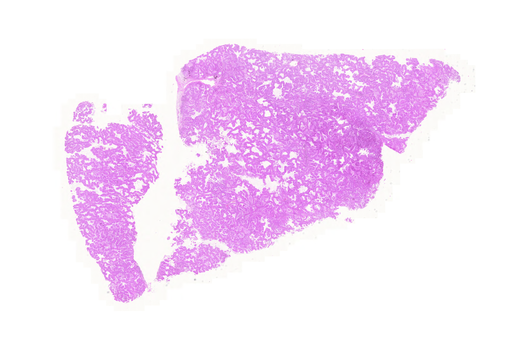}} &
\raisebox{-0.5\height}{\includegraphics[width=\hcolw]{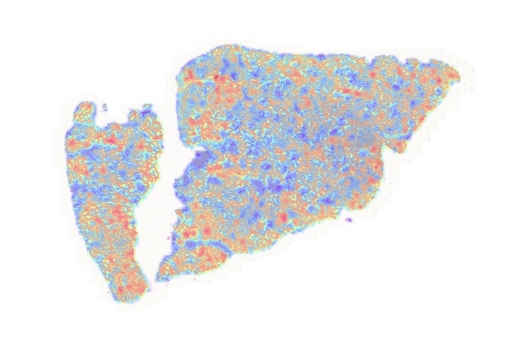}} &
\raisebox{-0.5\height}{\includegraphics[width=\hcolw]{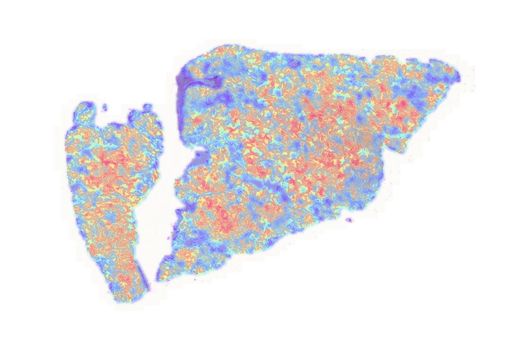}} \\[1pt]
\rotatebox[origin=c]{90}{\small\textbf{UCEC}} &
\raisebox{-0.5\height}{\includegraphics[width=\hcolw]{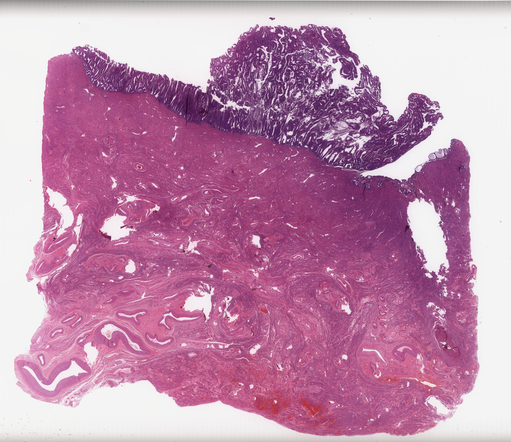}} &
\raisebox{-0.5\height}{\includegraphics[width=\hcolw]{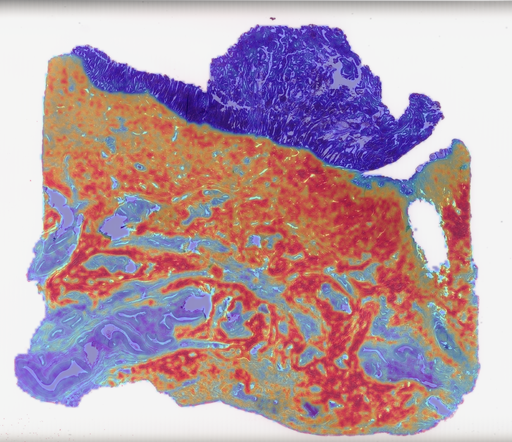}} &
\raisebox{-0.5\height}{\includegraphics[width=\hcolw]{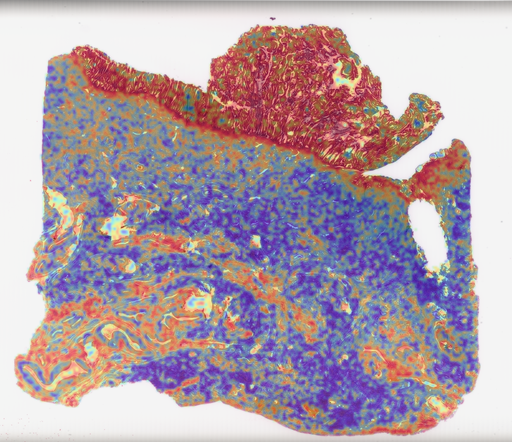}} \\[1pt]
\end{tabular}
\end{minipage}\hfill
\begin{minipage}[t]{0.28\linewidth}
\vspace{0pt}
\caption[Attention heatmaps]{\raggedright ABMIL attention without and with MFE-MIL (UNI; blue = low, red = high).
On CAMELYON16 the baseline attends to a tissue-fold artifact; MFE-MIL attends
to the tumour. TCGA-LUAD and UCEC are where the pathologist saw the clearest
improvement; all cohorts (including exceptions) and CLAM heads are in the
supplementary.}
\label{fig:heatmaps_main}
\end{minipage}
\end{figure*}

For the qualitative study, a single board-certified pathologist (unblinded, using a qualitative rubric; not a quantitative validation) reviewed 10 randomly selected WSIs
from each of the five TCGA survival cohorts (KIRC, KIRP, LUAD, STAD, UCEC) together with their attention maps; \cref{fig:heatmaps_main} shows examples (all cohorts and aggregators in the supplementary). The pathologist assessed how well attention focused on the lesion, whether the
high-attention areas within the lesion correlated with known pathology features, and how much
attention fell on background regions of low pathological value. Attention maps varied with both
task and dataset. Overall, MFE-MIL placed higher attention on lesion regions and showed better
visual agreement with pathologist-identified tissue features. This was most evident on the
TCGA-LUAD and UCEC cohorts; KIRC and STAD were the exceptions, while on KIRP the two models
produced similar maps.

\noindent\textbf{Quantitative attention localization.}
We further score attention against the
pixel-level tumour annotations on the CAMELYON16 test set
(\cref{tab:localization}).
Treating each patch's attention weight as a tumour score, MFE-MIL improves
localization AUROC for all three aggregators and pointing-game accuracy for ABMIL and CLAM-MB (unchanged for CLAM-SB), quantifying the improved tumour localization seen
in \cref{fig:heatmaps_main}.
A paired Wilcoxon signed-rank test over the slides with a defined patch-level
AUROC confirms the gain is significant for ABMIL
($p\!=\!4.1\!\times\!10^{-5}$) and CLAM-SB
($p\!=\!0.003$); the smaller CLAM-MB gain does not reach
significance ($p\!=\!0.056$).
This pattern mirrors the larger gains for weaker baselines: the gain is
largest and most reliable for the weakest-localizing baseline (ABMIL) and
smallest and not significant for the best-localizing one (CLAM-MB), consistent with the variance-suppression mechanism.
Top-50 attention mass is essentially unchanged,
indicating that MFE-MIL improves the \emph{ranking} of tumour patches rather
than concentrating additional total attention mass on them, consistent with
the class-selective entropy sharpening in \cref{tab:attn_entropy}.
A supplementary t-SNE figure further shows that MFE-MIL induces class-separable structure in the latent space.

\begin{table}[!tbp]
\centering
\scriptsize
\caption{Attention localization against CAMELYON16 tumour annotations
(UNI; the 48 of the 49 tumour test slides with usable pixel-level annotations).
AUROC: patch-level attention weight vs.\ annotation label;
Pointing: fraction of slides whose top-attended patch lies inside the
annotation; Top-50: fraction of top-50 attention mass inside the annotation.
Better value of each MIL\,/\,+MFE pair in \textbf{bold}.
Per-slide AUROC used for the paired significance test (text) is defined on
$n\!=\!43$ of these slides.}
\label{tab:localization}
\setlength{\tabcolsep}{5pt}
\begin{tabular}{@{}ll ccc@{}}
\toprule
Aggregator & Variant & AUROC & Pointing & Top-50 \\
\midrule
ABMIL   & MIL   & 0.879 & 0.688 & 0.598 \\
        & +MFE  & \textbf{0.942} & \textbf{0.812} & \textbf{0.605} \\[2pt]
CLAM-SB & MIL   & 0.893 & 0.750 & \textbf{0.598} \\
        & +MFE  & \textbf{0.921} & 0.750 & 0.592 \\[2pt]
CLAM-MB & MIL   & 0.902 & 0.708 & 0.593 \\
        & +MFE  & \textbf{0.913} & \textbf{0.729} & \textbf{0.600} \\
\bottomrule
\end{tabular}
\end{table}

\subsection{Geometric Analysis}
\label{sec:analysis}
We probe the adapter output on 128 held-out CAMELYON16 and 1031 held-out PANDA
WSIs, under classification alone versus with masked reconstruction.
Unlike the coordinate-free training grid, this analysis takes neighbours
directly from the pixel coordinates stored during patch extraction (patch step
$256$), so the neighbour sets below are genuinely spatial.
For each patch we compare cosine similarity to its 4-connected spatial
neighbours (adj) against similarity to 2000 randomly sampled distant patches
(rand); the gap adj$-$rand compares adjacent with random-pair similarity.

Together with the coherence ratio, three further metrics are reported:
normalised ABMIL attention entropy (\cref{tab:attn_entropy}), plus top-$k$
attention coherence and effective rank of the adapter output matrix (both in
the supplementary). All four agree that classification drives most of the compression; differences between masking variants are small.
Frozen UNI features start at a spatial coherence ratio of $1.952\times$. We contrast three conditions all with 2-layer MLP:
Baseline (classification-only adapter; ``MIL, 2'' in \cref{tab:ablation}), Rand MFE (single-token random masking; ``Single-token''), and Win MFE (window masking, our full model; ``Window (ours)''); results are in
\cref{tab:cam16_class}.

All three conditions dramatically compress within-slide variance. The ratio drops from $1.952\times$ to
$1.07$--$1.10\times$ regardless of masking strategy, which confirms that classification alone is the dominant driver of this semantic compression, i.e., the reduction of within-slide feature variance.
Across all masking variants the coherence gap falls in a narrow band with no reliable ordering by spatial arrangement or grouping: shuffling the packed order, using true-coordinate windows, 1D contiguous masks, distance-matched scattered groups, and single-token random masking all give comparable gaps under matched controls (supplementary), given the large split-to-split variation. Win MFE's larger CAMELYON16 gap comes mainly from lower random-pair similarity ($0.874$ versus $0.904$ for Baseline), while adjacent-pair similarity also drops ($0.958$ versus $0.968$), so it reflects greater global dispersion rather than closer neighbours. The gap therefore reflects the masked-reconstruction objective acting as a representation regularizer, rather than preservation of tissue-level spatial neighbourhoods. The larger window-versus-single-token gap on CAMELYON16 (\cref{tab:cam16_class}) is within this variation and is not reproduced on PANDA in the matched controls (0.058 versus 0.057; supplementary); in all cases the effect is modest and dataset-dependent, clearer on CAMELYON16 than on PANDA, where the gap barely moves even though downstream AUC still improves.

\begin{table}[!tbp]
\centering
\scriptsize
\caption{Spatial coherence on CAMELYON16 (UNI, 128 test WSIs: 49 tumour, 79
normal; one checkpoint per condition, split 0, seed 1; 4-connected neighbours
from patch coordinates): overall and class-conditional. Gap = adjacent minus random-pair similarity. Bold: highest
ratio and gap among the adapted conditions.
\textit{Left}: Frozen UNI and Baseline. \textit{Right}: Rand MFE and Win MFE.}
\label{tab:cam16_class}
\setlength{\tabcolsep}{3pt}
\begin{minipage}[t]{0.49\linewidth}
\centering
\begin{tabular}{@{}llcccc@{}}
\toprule
Condition & Class & adj & rand & ratio & gap \\
\midrule
Frozen UNI & overall & 0.654 & 0.335 & $1.952\times$ & 0.319 \\
           & tumor  & 0.655 & 0.328 & $1.997\times$ & 0.327 \\
           & normal & 0.653 & 0.339 & $1.925\times$ & 0.314 \\
\midrule
Baseline   & overall & 0.968 & 0.904 & $1.071\times$ & 0.064 \\
           & tumor  & 0.962 & 0.881 & $1.092\times$ & 0.081 \\
           & normal & 0.972 & 0.919 & $1.058\times$ & 0.053 \\
\bottomrule
\end{tabular}
\end{minipage}
\hfill
\begin{minipage}[t]{0.49\linewidth}
\centering
\begin{tabular}{@{}llcccc@{}}
\toprule
Condition & Class & adj & rand & ratio & gap \\
\midrule
Rand MFE   & overall & 0.975 & 0.911 & $1.070\times$ & 0.064 \\
           & tumor  & 0.971 & 0.896 & $1.084\times$ & 0.075 \\
           & normal & 0.977 & 0.920 & $1.062\times$ & 0.057 \\
\midrule
Win MFE    & overall & 0.958 & 0.874 & $1.095\times$ & 0.083 \\
           & tumor  & 0.953 & 0.848 & $\mathbf{1.123\times}$ & $\mathbf{0.105}$ \\
           & normal & 0.961 & 0.891 & $1.078\times$ & 0.070 \\
\bottomrule
\end{tabular}
\end{minipage}
\end{table}

Class-conditionally, block masking widens the tumour gap more than single-token masking on CAMELYON16, within split variation; CONCH class-conditional results are in the supplementary.

\noindent\textbf{Attention entropy analysis.} Semantic compression might flatten attention; instead it sharpens class-selectively (\cref{tab:attn_entropy}): on CAMELYON16, tumour entropy falls to around half while normal remains high, with differences between masking variants small and within run-to-run variation. On PANDA, adaptation sharpens attention for Grades~1--5 while Grade~0 (benign) remains higher, consistent with benign slides lacking a single focal lesion.

\begin{table}[!tbp]
\centering
\scriptsize
\caption{Normalised attention entropy (ABMIL, UNI).
Lower = sharper attention. \textit{Left}: CAMELYON16 class-conditional.
\textit{Right}: PANDA per Gleason grade.}
\label{tab:attn_entropy}
\setlength{\tabcolsep}{4pt}
\begin{minipage}[t]{0.44\linewidth}
\centering
\begin{tabular}{@{}lccc@{}}
\toprule
Condition & Overall & Normal & Tumor \\
\midrule
Frozen UNI & 0.985 & 0.985 & 0.986 \\
Baseline   & 0.814 & 0.981 & 0.544 \\
Rand MFE   & 0.794 & 0.984 & 0.488 \\
Win MFE    & 0.791 & 0.941 & 0.548 \\
\bottomrule
\end{tabular}
\end{minipage}
\hfill
\begin{minipage}[t]{0.54\linewidth}
\centering
\begin{tabular}{@{}lcccc@{}}
\toprule
Grade & Frozen & Baseline & Rand MFE & Win MFE \\
\midrule
G0     & 0.986 & 0.921 & 0.797 & 0.893 \\
G1              & 0.990 & 0.719 & 0.695 & 0.682 \\
G2              & 0.988 & 0.738 & 0.708 & 0.715 \\
G3              & 0.987 & 0.748 & 0.740 & 0.738 \\
G4              & 0.988 & 0.716 & 0.720 & 0.712 \\
G5  & 0.987 & 0.735 & 0.729 & 0.736 \\
\bottomrule
\end{tabular}
\end{minipage}
\end{table}

Finally, top-$k$ attention coherence (cosine similarity among the top-50 vs.\ bottom-50 attended patches) tells the same story (supplementary): on CAMELYON16 MFE variants widen the top$-$bottom gap, with lower similarity between attended and non-attended patches, so the two groups are pushed apart rather than homogenised; on PANDA the gap is essentially unchanged.
Together, the four metrics show that classification drives most of the compression and that masked reconstruction adds a modest regularizing effect whose size does not reliably depend on how the mask is arranged.
The compression direction generalises across encoders: CONCH features show the
same overall adj/rand ratio reduction as UNI, with benign tissue (Grade~0)
compressing the most (supplementary), supporting the encoder-agnostic claim.

\section{Conclusion}
\label{sec:conclusion}
MFE-MIL trains a lightweight MLP adapter jointly with a window-based masked
reconstruction branch and a MIL classification head.
The classification signal compresses within-slide patch variance; window-based masked reconstruction then further regularizes the adapted features.
Across CAMELYON16/17, PANDA, and TCGA-BRCA it improves ACC/F1 in nearly all settings,
outperforms CAMIL without using coordinate graphs at inference, and achieves
higher AUC than 2DMamba on three of four datasets (UNI). On five TCGA survival cohorts
it improves the average C-index for every aggregator tested.
Coherence, attention-entropy, and effective-rank analyses indicate that
classification drives most of the compression, with reconstruction adding a
modest regularizing effect. Limitations and ethical considerations are discussed
in the supplementary.
\bibliographystyle{splncs04}
\bibliography{main}

@String(CVPR  = {IEEE Conf. Comput. Vis. Pattern Recog.})

@String(ICCV  = {Int. Conf. Comput. Vis.})

@String(ECCV  = {Eur. Conf. Comput. Vis.})

@String(ICML  = {Int. Conf. Mach. Learn.})

@String(ICLR  = {Int. Conf. Learn. Represent.})

@String(CVPR  = {CVPR})

@String(ICCV  = {ICCV})

@String(ECCV  = {ECCV})

@String(ICML  = {ICML})

@String(ICLR  = {ICLR})

@misc{tcga,
  title        = {The Cancer Genome Atlas Program ({TCGA})},
  howpublished = {\url{https://www.cancer.gov/ccg/research/genome-sequencing/tcga}},
  year         = {2025}
}

@article{alagha2026atlaspatch,
  author  = {Alagha, A. and Leclerc, C. and Kotp, Y. and Metwally, O. and
             Moras, C. and Rentopoulos, P. and Rostami, G. and Nguyen, B.N. and
             Baig, J. and Khellaf, A. and Trinh, V.Q.-H. and Mizouni, R. and
             Otrok, H. and Bentahar, J. and Hosseini, M.S.},
  title   = {{AtlasPatch}: An efficient and scalable tool for whole slide image
             preprocessing in computational pathology},
  journal = {arXiv preprint arXiv:2602.03998},
  year    = {2026}
}

@article{bandi2019camelyon17,
  author  = {Bandi, P. and Geessink, O. and Manson, Q. and others},
  title   = {From detection of individual metastases to classification of lymph
             node status at the patient level: The {CAMELYON17} challenge},
  journal = {IEEE Transactions on Medical Imaging},
  volume  = {38},
  number  = {2},
  pages   = {550--560},
  year    = {2019}
}

@article{bejnordi2017camelyon16,
  author  = {Bejnordi, B.E. and Veta, M. and Van Diest, P.J. and others},
  title   = {Diagnostic assessment of deep learning algorithms for detection of
             lymph node metastases in women with breast cancer},
  journal = {JAMA},
  volume  = {318},
  number  = {22},
  pages   = {2199--2210},
  year    = {2017}
}

@article{bulten2022panda,
  author  = {Bulten, W. and Kartasalo, K. and Chen, P.H.C. and others},
  title   = {Artificial intelligence for diagnosis and {Gleason} grading of
             prostate cancer: the {PANDA} challenge},
  journal = {Nature Medicine},
  volume  = {28},
  pages   = {154--163},
  year    = {2022}
}

@article{chen2024uni,
  author  = {Chen, R.J. and Ding, T. and Lu, M.Y. and others},
  title   = {Towards a general-purpose foundation model for computational
             pathology},
  journal = {Nature Medicine},
  volume  = {30},
  pages   = {850--862},
  year    = {2024}
}

@inproceedings{chen2021mcat,
  author    = {Chen, R.J. and Lu, M.Y. and Weng, W.-H. and Chen, T.Y. and
               Williamson, D.F. and Manz, T. and Shady, M. and Mahmood, F.},
  title     = {Multimodal co-attention transformer for survival prediction in
               gigapixel whole slide images},
  booktitle = ICCV,
  pages     = {4015--4025},
  year      = {2021}
}

@inproceedings{jaume2024survpath,
  author    = {Jaume, G. and Vaidya, A. and Chen, R.J. and Williamson, D.F. and
               Liang, P.P. and Mahmood, F.},
  title     = {Modeling dense multimodal interactions between biological pathways
               and histology for survival prediction},
  booktitle = CVPR,
  year      = {2024}
}

@article{cifci2023ai,
  author  = {Cifci, D. and Veldhuizen, G.P. and Foersch, S. and Kather, J.N.},
  title   = {{AI} in computational pathology of cancer: Improving diagnostic
             workflows and clinical outcomes?},
  journal = {Annual Review of Cancer Biology},
  volume  = {7},
  pages   = {57--71},
  year    = {2023}
}

@article{ding2025titan,
  author  = {Ding, T. and Wagner, S.J. and Song, A.H. and others},
  title   = {A multimodal whole-slide foundation model for pathology},
  journal = {Nature Medicine},
  year    = {2025},
  doi     = {10.1038/s41591-025-03982-3}
}

@inproceedings{fang2024sammil,
  author    = {Fang, H. and Huang, S. and Tang, W. and others},
  title     = {{SAM-MIL}: A spatial contextual aware multiple instance learning
               approach for whole slide image classification},
  booktitle = {Proceedings of the 32nd ACM International Conference on Multimedia},
  pages     = {6083--6092},
  year      = {2024}
}

@inproceedings{fourkioti2024camil,
  author  = {Fourkioti, O. and De Vries, M. and Bakal, C.},
  title   = {{CAMIL}: Context-aware multiple instance learning for cancer
             detection and subtyping in whole slide images},
  booktitle = ICLR,
  year    = {2024}
}

@article{han2024peft,
  author  = {Han, Z. and Gao, C. and Liu, J. and others},
  title   = {Parameter-efficient fine-tuning for large models: A comprehensive
             survey},
  journal = {Transactions on Machine Learning Research},
  year    = {2024}
}

@inproceedings{he2022mae,
  author    = {He, K. and Chen, X. and Xie, S. and others},
  title     = {Masked autoencoders are scalable vision learners},
  booktitle = CVPR,
  pages     = {16000--16009},
  year      = {2022}
}

@article{hosseini2024computational,
  author  = {Hosseini, M.S. and Bejnordi, B.E. and Trinh, V.Q.H. and others},
  title   = {Computational pathology: A survey review and the way forward},
  journal = {Journal of Pathology Informatics},
  volume  = {15},
  pages   = {100357},
  year    = {2024}
}

@inproceedings{ilse2018attention,
  author    = {Ilse, M. and Tomczak, J. and Welling, M.},
  title     = {Attention-based deep multiple instance learning},
  booktitle = {International Conference on Machine Learning (ICML)},
  pages     = {2127--2136},
  year      = {2018}
}

@inproceedings{kang2023,
  author    = {Kang, M. and Song, H. and Park, S. and others},
  title     = {Benchmarking self-supervised learning on diverse pathology datasets},
  booktitle = CVPR,
  pages     = {3344--3354},
  year      = {2023}
}

@inproceedings{kirillov2023sam,
  author  = {Kirillov, A. and Mintun, E. and Ravi, N. and others},
  title   = {Segment anything},
  booktitle = ICCV,
  pages   = {4015--4026},
  year    = {2023}
}

@inproceedings{li2021dsmil,
  author    = {Li, B. and Li, Y. and Eliceiri, K.W.},
  title     = {Dual-stream multiple instance learning network for whole slide
               image classification with self-supervised contrastive learning},
  booktitle = CVPR,
  pages     = {14318--14328},
  year      = {2021}
}

@article{lu2021data,
  author    = {Lu, M.Y. and Williamson, D.F. and Chen, T.Y. and others},
  title     = {Data-efficient and weakly supervised computational pathology on
               whole-slide images},
  journal   = {Nature Biomedical Engineering},
  volume    = {5},
  number    = {6},
  pages     = {555--570},
  year      = {2021}
}

@article{lu2024conch,
  author  = {Lu, M.Y. and Chen, B. and Williamson, D.F. and others},
  title   = {A visual-language foundation model for computational pathology},
  journal = {Nature Medicine},
  volume  = {30},
  pages   = {863--874},
  year    = {2024}
}

@inproceedings{lu2023multimodal,
  author    = {Lu, M. and Wang, T. and Xia, Y.},
  title     = {Multi-modal pathological pre-training via masked autoencoders for
               breast cancer diagnosis},
  booktitle = {MICCAI},
  pages     = {457--466},
  year      = {2023}
}

@article{mammadov2025selfsupervision,
  author  = {Mammadov, A. and Le Folgoc, L. and Adam, J. and others},
  title   = {Self-supervision enhances instance-based multiple instance learning
             methods in digital pathology: A benchmark study},
  journal = {Journal of Medical Imaging},
  volume  = {12},
  number  = {6},
  pages   = {061404},
  year    = {2025}
}

@article{shaikovski2024prism,
  author  = {Shaikovski, G. and others},
  title   = {{PRISM}: A multi-modal generative foundation model for slide-level
             histopathology},
  journal = {arXiv preprint arXiv:2405.10254},
  year    = {2024}
}

@article{shao2021transmil,
  author  = {Shao, Z. and Bian, H. and Chen, Y. and others},
  title   = {{TransMIL}: Transformer based correlated multiple instance learning
             for whole slide image classification},
  journal = {Advances in Neural Information Processing Systems},
  volume  = {34},
  pages   = {2136--2147},
  year    = {2021}
}

@article{song2023artificial,
  author  = {Song, A.H. and Jaume, G. and Williamson, D.F. and others},
  title   = {Artificial intelligence for digital and computational pathology},
  journal = {Nature Reviews Bioengineering},
  volume  = {1},
  pages   = {930--949},
  year    = {2023}
}

@inproceedings{tang2023mhim,
  author    = {Tang, W. and Huang, S. and Zhang, X. and others},
  title     = {Multiple instance learning framework with masked hard instance
               mining for whole slide image classification},
  booktitle = ICCV,
  pages     = {4078--4087},
  year      = {2023}
}

@article{wang2019weakly,
  author  = {Wang, X. and Chen, H. and Gan, C. and others},
  title   = {Weakly supervised deep learning for whole slide lung cancer image
             analysis},
  journal = {IEEE Transactions on Cybernetics},
  volume  = {50},
  number  = {9},
  pages   = {3950--3962},
  year    = {2020}
}

@article{woelflein2023dasmil,
  author  = {W{\"o}lflein, G. and Magister, L.C. and Li{\`o}, P. and others},
  title   = {Deep multiple instance learning with distance-aware self-attention},
  journal = {arXiv preprint arXiv:2305.10552},
  year    = {2023}
}

@inproceedings{dosovitskiy2021image,
  author    = {Dosovitskiy, A. and Beyer, L. and Kolesnikov, A. and Weissenborn, D.
               and Zhai, X. and Unterthiner, T. and others},
  title     = {An Image is Worth 16x16 Words: Transformers for Image Recognition
               at Scale},
  booktitle = ICLR,
  year      = {2021}
}

@article{wu2024pancancer,
  author  = {Wu, K. and Jiang, Z. and Tang, K. and others},
  title   = {Pan-cancer histopathology {WSI} pre-training with position-aware
             masked autoencoder},
  journal = {IEEE Transactions on Medical Imaging},
  volume  = {44},
  number  = {4},
  pages   = {1610--1623},
  year    = {2025}
}

@article{xu2024gigapath,
  author  = {Xu, H. and others},
  title   = {A whole-slide foundation model for digital pathology from
             real-world data},
  journal = {Nature},
  volume  = {630},
  number  = {8015},
  pages   = {181--188},
  year    = {2024}
}

@inproceedings{zhang2022dtfd,
  author    = {Zhang, H. and Meng, Y. and Zhao, Y. and others},
  title     = {{DTFD-MIL}: Double-tier feature distillation multiple instance
               learning for histopathology whole slide image classification},
  booktitle = CVPR,
  pages     = {18802--18812},
  year      = {2022}
}

@article{zhang2025peft,
  author  = {Zhang, D. and Feng, T. and Xue, L. and others},
  title   = {Parameter-efficient fine-tuning for foundation models},
  journal = {arXiv preprint arXiv:2501.13787},
  year    = {2025}
}

@inproceedings{zhang2025_2dmamba,
  author    = {Zhang, J. and Nguyen, A.T. and Han, X. and others},
  title     = {{2DMamba}: Efficient state space model for image representation
               with applications on giga-pixel whole slide image classification},
  booktitle = CVPR,
  pages     = {3583--3592},
  year      = {2025}
}

@inproceedings{zhu2022setmil,
  author    = {Zhao, Y. and Lin, Z. and Sun, K. and Zhang, Y. and Huang, J. and
               Wang, L. and Yao, J.},
  title     = {{SETMIL}: Spatial encoding transformer-based multiple instance
               learning for pathological image analysis},
  booktitle = {MICCAI},
  pages     = {66--76},
  year      = {2022}
}

@inproceedings{zhang2024acmil,
    author = {Zhang, Yunlong and Li, Honglin and Sun, Yuxuan and Zheng, Sunyi and Zhu, Chenglu and Yang, Lin},
    title = {Attention-Challenging Multiple Instance Learning for Whole Slide Image Classification},
    booktitle = {European Conference on Computer Vision (ECCV)},
    year = {2024}
}

@inproceedings{li2024wikg,
    author = {Li, Jiawen and Chen, Yuxuan and Chu, Hongbo and Sun, Qiehe and Guan, Tian and Han, Anjia and He, Yonghong},
    title = {Dynamic Graph Representation with Knowledge-aware Attention for Histopathology Whole Slide Image Analysis},
    booktitle = {Proceedings of the IEEE/CVF Conference on Computer Vision and Pattern Recognition (CVPR)},
    pages = {11323--11332},
    year = {2024}
}

@inproceedings{expertcluster2026,
    author = {Papadopoulos, Kleanthis Marios and Giakoumoglou, Nikolaos and Floros, Andreas and Dragotti, Pier Luigi and Stathaki, Tania},
    title = {Expert Clustering and Knowledge Transfer for Whole Slide Image Classification},
    booktitle = {IEEE International Symposium on Biomedical Imaging (ISBI)},
    pages = {1--5},
    year = {2026},
    doi = {10.1109/ISBI61048.2026.11515371}
}

\end{document}


\title{Supplementary Material: Masked Feature Encoding for Large-Scale Whole Slide Image Representation}
\titlerunning{MFE-MIL: Supplementary Material}

\author{Haoyu He\inst{1}\orcidlink{0009-0004-6291-4913} \and
Basile Tessier-Cloutier\inst{3} \and
Yang Wang\inst{1,2} \and
Mahdi S.~Hosseini\inst{1,2,3}}
\authorrunning{H.~He et al.}

\institute{Department of Computer Science and Software Engineering (CSSE),
Concordia University, Montreal, Canada\\
\email{h\_haoy@live.concordia.ca}, \email{yang.wang@concordia.ca},
\email{mahdi.hosseini@concordia.ca} \and
Mila -- Quebec AI Institute, Montreal, Canada \and
McGill University, Montreal, Canada\\
\email{basile.tessiercloutier@mcgill.ca}}

\maketitle

\section{Training Algorithm}
\label{sec:supp_alg}

\Cref{alg:training} details the MFE-MIL training loop summarized in the main text.

\begin{algorithm}[!htbp]
\caption{MFE-MIL Training}
\label{alg:training}
\begin{algorithmic}[1]
\Require WSI bags $\{(\mathcal{X}^{(k)}, Y^{(k)})\}_{k=1}^{K}$, frozen encoder
         $f_\text{enc}$, masking ratio $r$, loss weight $\lambda_\text{rec}$
\Ensure Trained adapter $g_\theta$, decoder $f_\text{dec}$, MIL head ($f_\text{agg}$, $f_\text{class}$)
\For{each training bag $(\mathcal{X}, Y)$}
  \State $e_i \gets f_\text{enc}(x_i)$ for all $x_i \in \mathcal{X}$

  \State $h_i \gets g_\theta(e_i)$;\; form $H = [h_1,\ldots,h_N]^\top$ \Comment{MIL branch}
  \State $\bar{h}_i \gets g_\theta(\bar{e}_i)$ with $\bar{e}_i = e_i/\|e_i\|_2$ \Comment{reconstruction branch}

  \State Pack $\{\bar{h}_i\}$ into grid $\tilde{H} \in \mathbb{R}^{R\times C\times d}$
         in extraction order (row-major, as in the main paper)
  \State Sample adaptive window mask $M$,
         masking $\lfloor r \cdot RC \rfloor$ grid cells (padding included)

  \State $\tilde{H}' \gets \text{Mask}(\tilde{H}, M)$;\;
         $\hat{\tilde{H}} \gets f_\text{dec}(\tilde{H}', M, P)$

  \State $\mathcal{L}_\text{rec} \gets
         \frac{1}{|\mathcal{M}|\,D}\!\sum_{(i,j)\in\mathcal{M}}
         \|\bar{E}_{i,j,:} - \hat{\tilde{H}}_{i,j,:}\|_2^2$ (real patches only; $\bar{E}$: normalized frozen embeddings)

  \State $z \gets f_\text{agg}(H)$;\; $\hat{Y} \gets f_\text{class}(z)$;\;
         $\mathcal{L}_\text{MIL} \gets \text{CrossEntropy}(\hat{Y}, Y)$
  \State $\mathcal{L}_\text{total} \gets
         (1-\lambda_\text{rec})\,\mathcal{L}_\text{MIL}
         + \lambda_\text{rec}\,\mathcal{L}_\text{rec}$

  \State Backpropagate; update $g_\theta$, $f_\text{dec}$, $f_\text{agg}$, $f_\text{class}$;
         $f_\text{enc}$ remains \textbf{frozen}
\EndFor
\Statex \textbf{Inference:} discard $f_\text{dec}$;\;
       $e_i \xrightarrow{g_\theta} h_i \xrightarrow{f_\text{agg}} z
       \xrightarrow{f_\text{class}} \hat{Y}$
\end{algorithmic}
\end{algorithm}

\section{Full Numerical Results}
\label{sec:supp_results}

Tables~\ref{tab:full_uni}--\ref{tab:full_vits} report mean$\pm$std over
3 data splits $\times$ 5 seeds for all aggregators and all four encoders.
The spatially aware baselines CAMIL and 2DMamba and the masking-based MHIM-MIL
are reported for the UNI and CONCH encoders only.

\begin{figure}[!htbp]
  \centering
  \includegraphics[width=\linewidth]{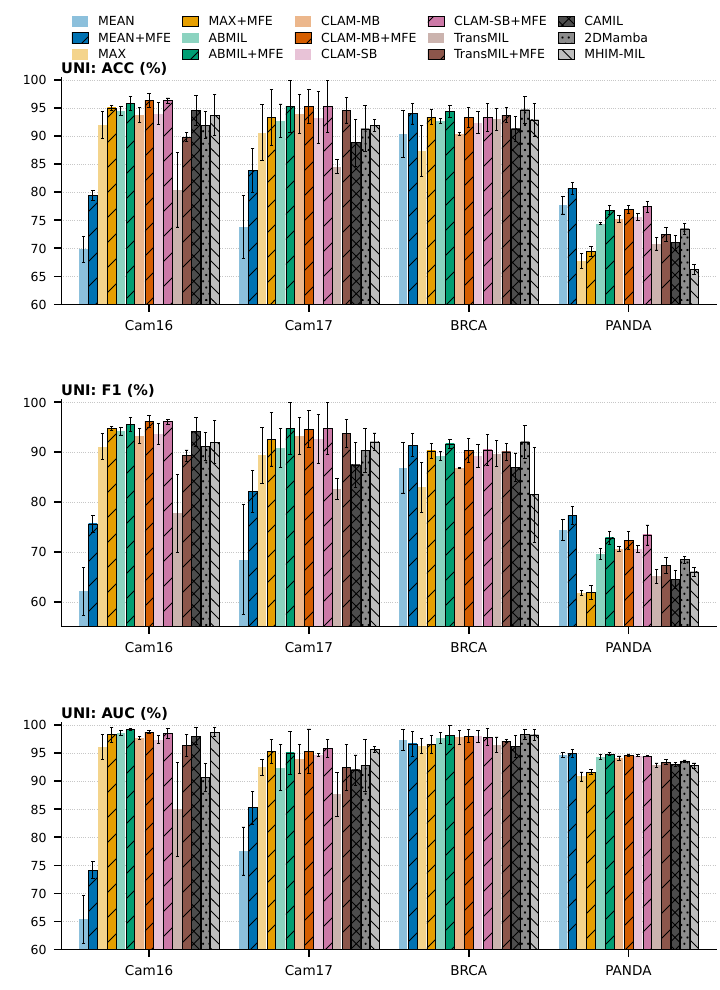}
  \caption{UNI counterpart of the main-text bar chart with all three metrics:
  ACC / F1 / AUC (\%) for all six MIL aggregators (lighter bar = base, darker
  hatched bar = +MFE-MIL) and the spatial baselines CAMIL, 2DMamba and MHIM-MIL
  (gray). Error bars show standard deviation over 3 splits $\times$ 5 seeds;
  Y-axes are truncated.}
  \label{fig:bars_uni_full}
\end{figure}

\begin{figure}[!htbp]
  \centering
  \includegraphics[width=\linewidth]{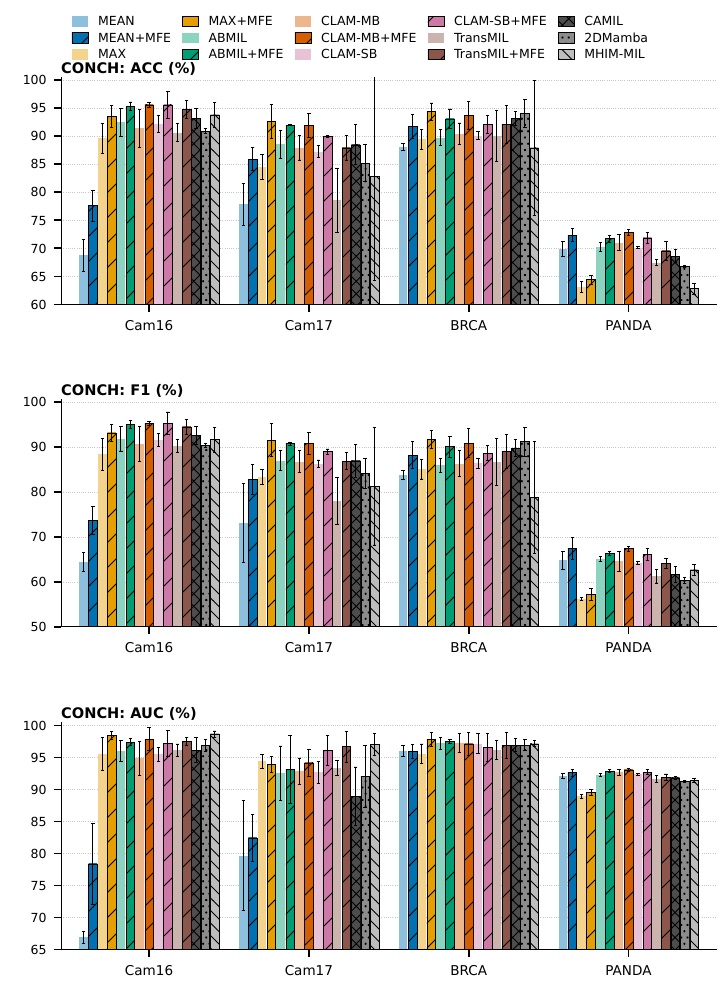}
  \caption{CONCH counterpart of the main-text bar chart: ACC / F1 / AUC (\%) for
  all six MIL aggregators (lighter bar = base, darker hatched bar = +MFE-MIL) and
  the spatial baselines CAMIL, 2DMamba and MHIM-MIL (gray). Error bars show
  standard deviation over 3 splits $\times$ 5 seeds; Y-axes are truncated.}
  \label{fig:bars_conch}
\end{figure}


\begin{table}[!htbp]
\centering
\scriptsize
\caption{Full results with \textbf{UNI} encoder.
ACC / F1 / AUC (\%), mean$_{\pm\text{std}}$, 3 splits $\times$ 5 seeds.
\textbf{Adapter (cls.\ only)} rows isolate the classification-trained adapter
without the reconstruction branch (frozen $\to$ adapter $\to$ +MFE
decomposition).}
\label{tab:full_uni}
\setlength{\tabcolsep}{3pt}
\resizebox{\linewidth}{!}{%
\begin{tabular}{@{}l ccc ccc ccc ccc@{}}
\toprule
& \multicolumn{3}{c}{\textbf{CAMELYON16}}
& \multicolumn{3}{c}{\textbf{CAMELYON17}}
& \multicolumn{3}{c}{\textbf{TCGA-BRCA}}
& \multicolumn{3}{c}{\textbf{PANDA}} \\
\cmidrule(lr){2-4}\cmidrule(lr){5-7}\cmidrule(lr){8-10}\cmidrule(lr){11-13}
Method & ACC & F1 & AUC & ACC & F1 & AUC & ACC & F1 & AUC & ACC & F1 & AUC \\
\midrule
MEAN
  & \ms{69.79}{2.25} & \ms{62.14}{4.83} & \ms{65.37}{4.30}
  & \ms{73.78}{5.66} & \ms{68.51}{11.06} & \ms{77.57}{4.28}
  & \ms{90.36}{4.19} & \ms{86.86}{5.02} & \ms{97.30}{1.86}
  & \ms{77.66}{1.54} & \ms{74.41}{2.10} & \ms{94.68}{0.50} \\
Adapter (cls.\ only)
  & \ms{78.33}{1.97} & \ms{74.91}{2.03} & \ms{73.88}{4.04}
  & \ms{77.85}{7.14} & \ms{76.06}{6.76} & \ms{82.42}{4.99}
  & \ms{91.81}{3.26} & \ms{88.52}{3.85} & \ms{96.05}{1.84}
  & \ms{78.49}{1.57} & \ms{74.65}{2.16} & \ms{94.27}{0.63} \\
MFE-MEAN
  & \ms{79.43}{0.90} & \ms{75.65}{1.71} & \ms{74.14}{1.49}
  & \ms{83.92}{3.88} & \ms{82.15}{4.20} & \ms{85.28}{2.90}
  & \ms{94.01}{1.84} & \ms{91.40}{2.28} & \ms{96.68}{2.21}
  & \ms{80.67}{1.08} & \ms{77.34}{1.77} & \ms{94.93}{0.68} \\
\midrule
MAX
  & \ms{91.93}{2.39} & \ms{91.12}{2.66} & \ms{96.15}{2.27}
  & \ms{90.64}{4.99} & \ms{89.41}{5.63} & \ms{92.46}{1.44}
  & \ms{87.32}{4.52} & \ms{82.95}{4.93} & \ms{96.31}{1.36}
  & \ms{67.77}{1.37} & \ms{61.75}{0.49} & \ms{90.85}{0.80} \\
Adapter (cls.\ only)
  & \ms{92.60}{4.12} & \ms{92.26}{4.07} & \ms{98.62}{0.70}
  & \ms{93.04}{3.97} & \ms{92.31}{4.24} & \ms{95.53}{2.32}
  & \ms{91.33}{3.27} & \ms{86.73}{3.54} & \ms{96.23}{1.56}
  & \ms{68.19}{2.12} & \ms{61.40}{1.95} & \ms{91.10}{0.42} \\
MFE-MAX
  & \ms{95.05}{0.45} & \ms{94.77}{0.47} & \ms{98.31}{1.32}
  & \ms{93.32}{5.02} & \ms{92.61}{5.36} & \ms{95.31}{2.12}
  & \ms{93.41}{1.29} & \ms{90.27}{1.49} & \ms{96.59}{1.63}
  & \ms{69.41}{0.94} & \ms{61.88}{1.43} & \ms{91.62}{0.48} \\
\midrule
ABMIL
  & \ms{94.53}{0.78} & \ms{94.21}{0.76} & \ms{98.60}{0.45}
  & \ms{92.76}{2.98} & \ms{90.89}{3.88} & \ms{92.45}{4.09}
  & \ms{92.71}{0.49} & \ms{89.30}{0.91} & \ms{97.69}{1.02}
  & \ms{74.39}{0.17} & \ms{69.65}{1.11} & \ms{94.34}{0.40} \\
Adapter (cls.\ only)
  & \ms{94.79}{3.82} & \ms{94.39}{4.20} & \ms{97.04}{2.32}
  & \ms{92.11}{5.51} & \ms{91.35}{5.70} & \ms{94.58}{3.09}
  & \ms{93.25}{2.34} & \ms{90.66}{2.49} & \ms{97.63}{1.93}
  & \ms{76.14}{1.25} & \ms{71.74}{2.02} & \ms{94.23}{0.36} \\
MFE-ABMIL
  & \ms{95.83}{1.20} & \ms{95.53}{1.36} & \ms{99.25}{0.19}
  & \ms{95.32}{4.61} & \ms{94.71}{5.21} & \ms{95.05}{3.78}
  & \ms{94.38}{1.04} & \ms{91.66}{0.92} & \ms{98.19}{1.68}
  & \ms{76.78}{0.78} & \ms{72.84}{1.34} & \ms{94.85}{0.30} \\
\midrule
CLAM-MB
  & \ms{93.75}{1.35} & \ms{93.27}{1.46} & \ms{97.72}{0.34}
  & \ms{93.97}{3.44} & \ms{93.30}{3.72} & \ms{93.93}{2.58}
  & \ms{90.40}{0.20} & \ms{86.84}{0.09} & \ms{97.84}{1.25}
  & \ms{75.20}{0.60} & \ms{70.57}{0.53} & \ms{94.13}{0.38} \\
Adapter (cls.\ only)
  & \ms{95.21}{1.84} & \ms{94.89}{1.94} & \ms{97.63}{1.64}
  & \ms{92.91}{5.05} & \ms{92.07}{5.21} & \ms{94.33}{4.73}
  & \ms{92.14}{1.76} & \ms{89.18}{1.65} & \ms{97.63}{1.73}
  & \ms{75.38}{1.50} & \ms{70.94}{2.07} & \ms{94.24}{0.44} \\
MFE-CLAMB
  & \ms{96.35}{1.20} & \ms{96.13}{1.25} & \ms{98.73}{0.27}
  & \ms{95.32}{3.04} & \ms{94.59}{3.70} & \ms{95.36}{3.96}
  & \ms{93.34}{1.74} & \ms{90.36}{2.43} & \ms{98.00}{1.16}
  & \ms{76.92}{0.68} & \ms{72.34}{1.84} & \ms{94.67}{0.19} \\
\midrule
CLAM-SB
  & \ms{94.01}{1.96} & \ms{93.64}{2.03} & \ms{97.42}{0.69}
  & \ms{93.31}{4.60} & \ms{92.63}{4.89} & \ms{94.68}{0.21}
  & \ms{92.39}{1.96} & \ms{89.21}{2.28} & \ms{97.99}{1.12}
  & \ms{75.62}{0.65} & \ms{70.68}{0.67} & \ms{94.57}{0.17} \\
Adapter (cls.\ only)
  & \ms{94.38}{4.24} & \ms{94.09}{4.25} & \ms{97.71}{1.86}
  & \ms{91.03}{7.48} & \ms{90.24}{7.44} & \ms{94.29}{3.58}
  & \ms{92.46}{3.31} & \ms{90.25}{3.87} & \ms{97.72}{2.04}
  & \ms{75.77}{1.51} & \ms{71.40}{1.98} & \ms{94.21}{0.52} \\
MFE-CLAMSB
  & \ms{96.35}{0.46} & \ms{96.12}{0.48} & \ms{98.48}{0.88}
  & \ms{95.32}{4.61} & \ms{94.71}{5.21} & \ms{95.85}{1.58}
  & \ms{93.36}{2.50} & \ms{90.49}{3.15} & \ms{97.88}{1.58}
  & \ms{77.43}{0.98} & \ms{73.43}{2.00} & \ms{94.50}{0.15} \\
\midrule
TransMIL
  & \ms{80.47}{6.67} & \ms{77.75}{7.76} & \ms{84.98}{8.32}
  & \ms{84.56}{1.26} & \ms{82.64}{2.14} & \ms{87.66}{3.97}
  & \ms{93.04}{1.97} & \ms{89.70}{2.60} & \ms{96.48}{1.41}
  & \ms{70.84}{1.14} & \ms{65.16}{1.40} & \ms{92.86}{0.40} \\
MFE-TransMIL
  & \ms{89.84}{0.78} & \ms{89.34}{1.11} & \ms{96.43}{1.98}
  & \ms{94.64}{2.29} & \ms{93.74}{2.83} & \ms{92.44}{4.05}
  & \ms{93.73}{1.34} & \ms{90.06}{1.67} & \ms{97.17}{0.22}
  & \ms{72.52}{1.26} & \ms{67.27}{1.57} & \ms{93.43}{0.45} \\
\midrule
MHIM-MIL
  & \ms{93.76}{3.72} & \ms{92.01}{4.30} & \ms{98.74}{0.80}
  & \ms{91.91}{1.14} & \ms{92.07}{1.79} & \ms{95.67}{0.47}
  & \ms{92.89}{2.87} & \ms{81.50}{9.49} & \ms{98.28}{0.95}
  & \ms{66.22}{0.83} & \ms{66.01}{0.85} & \ms{92.79}{0.47} \\
2DMamba
  & \ms{91.93}{2.39} & \ms{91.13}{2.74} & \ms{90.73}{2.51}
  & \ms{91.31}{4.12} & \ms{90.34}{4.43} & \ms{92.80}{4.58}
  & \ms{94.70}{2.43} & \ms{92.06}{3.32} & \ms{98.31}{0.85}
  & \ms{73.46}{1.02} & \ms{68.48}{0.71} & \ms{93.58}{0.22} \\
CAMIL
  & \ms{94.53}{2.67} & \ms{94.13}{2.88} & \ms{98.06}{1.45}
  & \ms{88.85}{4.20} & \ms{87.44}{4.51} & \ms{92.01}{2.67}
  & \ms{91.30}{2.20} & \ms{86.98}{2.88} & \ms{96.24}{2.00}
  & \ms{71.01}{1.33} & \ms{64.47}{1.83} & \ms{92.97}{0.37} \\
\bottomrule
\end{tabular}
}
\end{table}

\begin{table}[!htbp]
\centering
\scriptsize
\caption{Full results with \textbf{CONCH} encoder.
ACC / F1 / AUC (\%), mean$_{\pm\text{std}}$, 3 splits $\times$ 5 seeds.}
\label{tab:full_conch}
\setlength{\tabcolsep}{3pt}
\resizebox{\linewidth}{!}{%
\begin{tabular}{@{}l ccc ccc ccc ccc@{}}
\toprule
& \multicolumn{3}{c}{\textbf{CAMELYON16}}
& \multicolumn{3}{c}{\textbf{CAMELYON17}}
& \multicolumn{3}{c}{\textbf{TCGA-BRCA}}
& \multicolumn{3}{c}{\textbf{PANDA}} \\
\cmidrule(lr){2-4}\cmidrule(lr){5-7}\cmidrule(lr){8-10}\cmidrule(lr){11-13}
Method & ACC & F1 & AUC & ACC & F1 & AUC & ACC & F1 & AUC & ACC & F1 & AUC \\
\midrule
MEAN
  & \ms{68.75}{2.81} & \ms{64.45}{2.09} & \ms{66.94}{0.94}
  & \ms{77.82}{3.77} & \ms{73.19}{8.78} & \ms{79.68}{8.62}
  & \ms{88.09}{0.62} & \ms{83.73}{1.03} & \ms{96.06}{0.84}
  & \ms{69.93}{1.37} & \ms{64.78}{2.00} & \ms{92.12}{0.41} \\
MFE-MEAN
  & \ms{77.60}{2.74} & \ms{73.70}{3.15} & \ms{78.38}{6.31}
  & \ms{85.90}{2.01} & \ms{82.80}{3.30} & \ms{82.44}{3.66}
  & \ms{91.74}{2.11} & \ms{88.21}{3.01} & \ms{95.95}{1.16}
  & \ms{72.36}{1.18} & \ms{67.47}{2.47} & \ms{92.64}{0.54} \\
\midrule
MAX
  & \ms{89.58}{2.74} & \ms{88.39}{3.61} & \ms{95.53}{2.55}
  & \ms{84.57}{2.23} & \ms{83.34}{1.68} & \ms{94.41}{1.13}
  & \ms{89.43}{1.79} & \ms{85.04}{2.13} & \ms{95.59}{1.53}
  & \ms{63.14}{1.02} & \ms{56.29}{0.30} & \ms{88.92}{0.29} \\
MFE-MAX
  & \ms{93.49}{1.96} & \ms{93.17}{1.93} & \ms{98.47}{0.57}
  & \ms{92.61}{3.07} & \ms{91.56}{3.75} & \ms{93.91}{1.26}
  & \ms{94.36}{1.55} & \ms{91.73}{1.99} & \ms{97.82}{1.10}
  & \ms{64.41}{0.83} & \ms{57.23}{1.36} & \ms{89.58}{0.50} \\
\midrule
ABMIL
  & \ms{92.45}{2.51} & \ms{91.87}{2.75} & \ms{96.01}{1.69}
  & \ms{88.57}{2.48} & \ms{87.00}{2.25} & \ms{92.57}{4.24}
  & \ms{89.73}{1.46} & \ms{85.96}{1.53} & \ms{97.22}{0.89}
  & \ms{70.22}{0.79} & \ms{65.05}{0.56} & \ms{92.28}{0.23} \\
MFE-ABMIL
  & \ms{95.31}{0.78} & \ms{94.99}{0.85} & \ms{97.41}{0.56}
  & \ms{91.95}{0.09} & \ms{90.76}{0.31} & \ms{93.12}{5.32}
  & \ms{93.08}{1.77} & \ms{90.06}{2.30} & \ms{97.55}{0.30}
  & \ms{71.71}{0.60} & \ms{66.32}{0.49} & \ms{92.84}{0.23} \\
\midrule
CLAM-MB
  & \ms{91.40}{3.41} & \ms{90.62}{3.91} & \ms{94.89}{2.69}
  & \ms{87.90}{2.15} & \ms{86.78}{2.42} & \ms{92.86}{2.07}
  & \ms{90.40}{1.95} & \ms{86.32}{2.85} & \ms{97.26}{1.50}
  & \ms{71.00}{1.46} & \ms{64.57}{2.22} & \ms{92.70}{0.48} \\
MFE-CLAMB
  & \ms{95.57}{0.45} & \ms{95.31}{0.47} & \ms{97.86}{1.82}
  & \ms{91.93}{2.10} & \ms{90.82}{2.43} & \ms{94.17}{2.15}
  & \ms{93.71}{2.43} & \ms{90.83}{3.21} & \ms{97.14}{1.80}
  & \ms{72.78}{0.54} & \ms{67.41}{0.52} & \ms{93.07}{0.24} \\
\midrule
CLAM-SB
  & \ms{92.19}{1.56} & \ms{91.57}{1.51} & \ms{95.48}{1.08}
  & \ms{87.25}{1.09} & \ms{86.25}{0.87} & \ms{92.70}{1.69}
  & \ms{90.08}{0.68} & \ms{86.34}{1.03} & \ms{97.16}{1.54}
  & \ms{70.13}{0.20} & \ms{64.29}{0.36} & \ms{92.38}{0.20} \\
MFE-CLAMSB
  & \ms{95.57}{2.38} & \ms{95.27}{2.53} & \ms{97.24}{2.06}
  & \ms{89.93}{0.12} & \ms{88.98}{0.60} & \ms{96.09}{2.31}
  & \ms{92.07}{1.56} & \ms{88.66}{1.68} & \ms{96.57}{2.24}
  & \ms{71.84}{0.93} & \ms{66.12}{1.41} & \ms{92.73}{0.35} \\
\midrule
TransMIL
  & \ms{90.63}{1.57} & \ms{90.27}{1.50} & \ms{96.17}{0.95}
  & \ms{78.54}{5.67} & \ms{77.94}{5.20} & \ms{93.42}{1.18}
  & \ms{90.03}{4.56} & \ms{86.66}{5.25} & \ms{96.19}{1.46}
  & \ms{67.44}{0.50} & \ms{61.32}{1.54} & \ms{91.68}{0.55} \\
MFE-TransMIL
  & \ms{94.79}{1.63} & \ms{94.48}{1.69} & \ms{97.56}{0.59}
  & \ms{87.90}{2.15} & \ms{86.87}{1.84} & \ms{96.71}{2.40}
  & \ms{92.11}{3.35} & \ms{89.08}{3.75} & \ms{96.88}{2.02}
  & \ms{69.54}{1.65} & \ms{64.12}{1.04} & \ms{91.94}{0.49} \\
\midrule
MHIM-MIL
  & \ms{93.78}{2.24} & \ms{91.64}{2.74} & \ms{98.68}{0.48}
  & \ms{82.85}{18.57} & \ms{81.25}{13.02} & \ms{97.02}{1.70}
  & \ms{87.92}{11.99} & \ms{78.73}{12.41} & \ms{97.15}{0.45}
  & \ms{62.83}{0.98} & \ms{62.71}{1.21} & \ms{91.44}{0.27} \\
2DMamba
  & \ms{90.89}{0.45} & \ms{90.36}{0.45} & \ms{96.87}{0.95}
  & \ms{85.21}{3.26} & \ms{84.12}{3.43} & \ms{92.05}{4.90}
  & \ms{94.06}{2.51} & \ms{91.22}{3.23} & \ms{96.94}{0.85}
  & \ms{66.76}{0.24} & \ms{60.33}{0.72} & \ms{91.27}{0.22} \\
CAMIL
  & \ms{93.12}{1.83} & \ms{92.56}{2.09} & \ms{96.14}{1.97}
  & \ms{88.46}{3.55} & \ms{86.93}{3.65} & \ms{88.97}{4.56}
  & \ms{93.13}{1.25} & \ms{89.78}{1.81} & \ms{96.92}{1.02}
  & \ms{68.61}{1.19} & \ms{61.66}{1.71} & \ms{91.82}{0.29} \\
\bottomrule
\end{tabular}
}
\end{table}

\begin{table}[!htbp]
\centering
\scriptsize
\caption{Full results with \textbf{ViT-L/16-IN21K} encoder.
ACC / F1 / AUC (\%), mean$_{\pm\text{std}}$, 3 splits $\times$ 5 seeds.
Baselines (CAMIL, 2DMamba, MHIM-MIL) are reported for UNI and CONCH only.}
\label{tab:full_vitl}
\setlength{\tabcolsep}{3pt}
\resizebox{\linewidth}{!}{%
\begin{tabular}{@{}l ccc ccc ccc ccc@{}}
\toprule
& \multicolumn{3}{c}{\textbf{CAMELYON16}}
& \multicolumn{3}{c}{\textbf{CAMELYON17}}
& \multicolumn{3}{c}{\textbf{TCGA-BRCA}}
& \multicolumn{3}{c}{\textbf{PANDA}} \\
\cmidrule(lr){2-4}\cmidrule(lr){5-7}\cmidrule(lr){8-10}\cmidrule(lr){11-13}
Method & ACC & F1 & AUC & ACC & F1 & AUC & ACC & F1 & AUC & ACC & F1 & AUC \\
\midrule
MEAN
  & \ms{60.94}{5.12} & \ms{49.11}{4.81} & \ms{54.94}{3.54}
  & \ms{61.69}{8.47} & \ms{59.11}{6.40} & \ms{68.86}{11.42}
  & \ms{87.03}{2.49} & \ms{82.86}{3.04} & \ms{92.57}{0.96}
  & \ms{61.56}{1.51} & \ms{57.15}{1.34} & \ms{88.89}{0.71} \\
MFE-MEAN
  & \ms{68.75}{4.36} & \ms{65.74}{2.86} & \ms{67.38}{2.72}
  & \ms{71.14}{1.03} & \ms{67.34}{0.84} & \ms{71.69}{7.92}
  & \ms{89.76}{0.76} & \ms{85.41}{0.56} & \ms{91.80}{5.14}
  & \ms{65.60}{1.81} & \ms{60.30}{3.42} & \ms{89.63}{0.67} \\
\midrule
MAX
  & \ms{72.66}{2.35} & \ms{70.25}{1.26} & \ms{75.75}{1.37}
  & \ms{67.76}{7.15} & \ms{61.43}{13.16} & \ms{70.35}{16.21}
  & \ms{76.16}{0.50} & \ms{66.46}{5.75} & \ms{79.09}{12.29}
  & \ms{56.06}{0.34} & \ms{50.11}{0.85} & \ms{85.75}{0.39} \\
MFE-MAX
  & \ms{76.30}{1.19} & \ms{73.70}{2.17} & \ms{77.73}{2.25}
  & \ms{77.17}{2.46} & \ms{74.89}{2.50} & \ms{80.53}{4.70}
  & \ms{84.39}{2.06} & \ms{74.90}{1.04} & \ms{82.72}{3.66}
  & \ms{57.45}{0.24} & \ms{51.33}{1.21} & \ms{86.43}{0.65} \\
\midrule
ABMIL
  & \ms{75.00}{2.70} & \ms{72.36}{3.36} & \ms{73.76}{3.21}
  & \ms{67.10}{1.55} & \ms{65.67}{1.71} & \ms{78.74}{3.04}
  & \ms{87.30}{4.37} & \ms{80.66}{6.09} & \ms{91.28}{3.17}
  & \ms{61.20}{1.21} & \ms{55.59}{1.08} & \ms{88.15}{0.93} \\
MFE-ABMIL
  & \ms{78.65}{1.20} & \ms{75.23}{1.55} & \ms{76.43}{6.87}
  & \ms{83.21}{1.37} & \ms{80.42}{3.01} & \ms{82.81}{5.43}
  & \ms{92.34}{1.77} & \ms{87.92}{3.41} & \ms{94.56}{3.82}
  & \ms{63.95}{0.44} & \ms{58.27}{1.32} & \ms{89.36}{0.35} \\
\midrule
CLAM-MB
  & \ms{73.70}{1.19} & \ms{70.67}{0.24} & \ms{75.99}{0.86}
  & \ms{76.49}{2.62} & \ms{73.14}{2.76} & \ms{78.36}{0.75}
  & \ms{87.30}{4.58} & \ms{79.90}{6.84} & \ms{91.32}{1.99}
  & \ms{60.26}{0.81} & \ms{53.45}{3.04} & \ms{87.34}{0.74} \\
MFE-CLAMB
  & \ms{77.35}{2.82} & \ms{74.19}{3.13} & \ms{78.72}{3.99}
  & \ms{83.85}{5.55} & \ms{81.41}{4.71} & \ms{82.60}{3.06}
  & \ms{90.01}{2.17} & \ms{83.34}{5.09} & \ms{92.94}{5.12}
  & \ms{63.66}{1.43} & \ms{57.53}{1.81} & \ms{89.43}{0.54} \\
\midrule
CLAM-SB
  & \ms{73.18}{1.97} & \ms{70.31}{1.94} & \ms{72.73}{2.86}
  & \ms{79.21}{2.10} & \ms{76.30}{1.48} & \ms{76.37}{1.51}
  & \ms{86.98}{4.06} & \ms{80.93}{4.81} & \ms{91.01}{7.26}
  & \ms{60.88}{0.82} & \ms{54.74}{1.78} & \ms{87.71}{0.95} \\
MFE-CLAMSB
  & \ms{77.08}{1.20} & \ms{73.36}{3.53} & \ms{77.14}{1.74}
  & \ms{84.56}{3.10} & \ms{80.26}{4.96} & \ms{80.83}{0.48}
  & \ms{90.63}{3.64} & \ms{85.74}{5.77} & \ms{93.36}{4.96}
  & \ms{62.82}{0.50} & \ms{57.22}{0.73} & \ms{89.20}{0.56} \\
\midrule
TransMIL
  & \ms{63.54}{2.39} & \ms{60.96}{2.32} & \ms{68.62}{3.81}
  & \ms{68.45}{1.38} & \ms{66.34}{2.53} & \ms{74.36}{1.16}
  & \ms{87.14}{2.17} & \ms{77.49}{4.21} & \ms{89.96}{2.40}
  & \ms{58.00}{0.50} & \ms{52.30}{0.22} & \ms{87.65}{0.25} \\
MFE-TransMIL
  & \ms{66.41}{3.91} & \ms{62.28}{5.23} & \ms{69.90}{4.55}
  & \ms{77.16}{3.35} & \ms{72.83}{4.04} & \ms{77.80}{9.21}
  & \ms{87.17}{3.15} & \ms{78.63}{9.32} & \ms{92.37}{3.10}
  & \ms{60.52}{0.73} & \ms{54.59}{0.56} & \ms{87.76}{0.83} \\

\bottomrule
\end{tabular}
}
\end{table}

\begin{table}[!htbp]
\centering
\scriptsize
\caption{Full results with \textbf{ViT-S/16-SSL} encoder.
ACC / F1 / AUC (\%), mean$_{\pm\text{std}}$, 3 splits $\times$ 5 seeds.
Baselines (CAMIL, 2DMamba, MHIM-MIL) are reported for UNI and CONCH only.}
\label{tab:full_vits}
\setlength{\tabcolsep}{3pt}
\resizebox{\linewidth}{!}{%
\begin{tabular}{@{}l ccc ccc ccc ccc@{}}
\toprule
& \multicolumn{3}{c}{\textbf{CAMELYON16}}
& \multicolumn{3}{c}{\textbf{CAMELYON17}}
& \multicolumn{3}{c}{\textbf{TCGA-BRCA}}
& \multicolumn{3}{c}{\textbf{PANDA}} \\
\cmidrule(lr){2-4}\cmidrule(lr){5-7}\cmidrule(lr){8-10}\cmidrule(lr){11-13}
Method & ACC & F1 & AUC & ACC & F1 & AUC & ACC & F1 & AUC & ACC & F1 & AUC \\
\midrule
MEAN
  & \ms{72.14}{2.74} & \ms{66.23}{5.38} & \ms{67.77}{3.99}
  & \ms{73.13}{3.39} & \ms{67.04}{5.39} & \ms{75.48}{8.20}
  & \ms{86.45}{1.72} & \ms{81.79}{1.62} & \ms{95.06}{2.37}
  & \ms{73.52}{1.08} & \ms{69.67}{1.16} & \ms{93.44}{0.49} \\
MFE-MEAN
  & \ms{73.96}{3.25} & \ms{69.13}{2.47} & \ms{73.98}{3.21}
  & \ms{81.18}{4.35} & \ms{77.96}{5.98} & \ms{81.62}{3.31}
  & \ms{92.41}{2.69} & \ms{89.04}{3.29} & \ms{95.36}{2.81}
  & \ms{76.04}{0.49} & \ms{72.04}{1.20} & \ms{93.50}{0.36} \\
\midrule
MAX
  & \ms{94.27}{1.19} & \ms{93.79}{1.34} & \ms{98.62}{0.83}
  & \ms{87.90}{2.15} & \ms{86.44}{2.35} & \ms{91.47}{2.52}
  & \ms{89.01}{3.66} & \ms{84.31}{4.68} & \ms{93.64}{1.16}
  & \ms{65.25}{2.05} & \ms{58.84}{2.44} & \ms{90.13}{0.89} \\
MFE-MAX
  & \ms{94.79}{0.45} & \ms{94.49}{0.36} & \ms{99.12}{0.33}
  & \ms{90.60}{1.22} & \ms{89.35}{0.98} & \ms{95.64}{1.33}
  & \ms{91.67}{3.85} & \ms{87.83}{5.38} & \ms{95.17}{2.61}
  & \ms{67.60}{0.20} & \ms{61.77}{0.33} & \ms{90.78}{0.49} \\
\midrule
ABMIL
  & \ms{89.06}{2.07} & \ms{88.55}{1.64} & \ms{96.20}{3.14}
  & \ms{88.60}{1.04} & \ms{87.23}{1.00} & \ms{93.25}{3.21}
  & \ms{92.04}{1.02} & \ms{88.68}{1.76} & \ms{96.92}{1.12}
  & \ms{72.39}{0.62} & \ms{67.40}{0.66} & \ms{93.44}{0.25} \\
MFE-ABMIL
  & \ms{91.93}{0.90} & \ms{91.58}{0.96} & \ms{97.41}{1.19}
  & \ms{93.31}{3.02} & \ms{92.64}{2.90} & \ms{92.96}{1.95}
  & \ms{94.03}{1.96} & \ms{90.82}{2.76} & \ms{97.86}{0.26}
  & \ms{75.66}{2.83} & \ms{71.47}{2.96} & \ms{94.39}{0.48} \\
\midrule
CLAM-MB
  & \ms{93.75}{0.78} & \ms{93.29}{0.68} & \ms{97.84}{0.64}
  & \ms{85.92}{3.40} & \ms{84.85}{3.27} & \ms{91.30}{0.68}
  & \ms{93.38}{0.32} & \ms{90.31}{0.24} & \ms{96.56}{2.46}
  & \ms{74.39}{1.18} & \ms{70.76}{1.32} & \ms{94.01}{0.48} \\
MFE-CLAMB
  & \ms{94.53}{0.78} & \ms{94.29}{0.80} & \ms{98.08}{1.01}
  & \ms{89.27}{3.04} & \ms{88.32}{3.09} & \ms{95.81}{3.01}
  & \ms{93.36}{0.74} & \ms{90.30}{0.76} & \ms{97.87}{0.32}
  & \ms{75.14}{1.12} & \ms{71.32}{1.62} & \ms{94.20}{0.52} \\
\midrule
CLAM-SB
  & \ms{93.75}{0.78} & \ms{93.19}{0.82} & \ms{97.37}{1.34}
  & \ms{89.25}{2.38} & \ms{87.79}{2.86} & \ms{94.59}{3.11}
  & \ms{91.38}{3.46} & \ms{86.94}{4.72} & \ms{97.40}{1.21}
  & \ms{73.62}{3.07} & \ms{69.38}{2.85} & \ms{93.78}{0.86} \\
MFE-CLAMSB
  & \ms{94.53}{0.78} & \ms{94.10}{0.85} & \ms{98.07}{0.65}
  & \ms{91.93}{2.10} & \ms{91.03}{2.42} & \ms{96.38}{2.25}
  & \ms{92.74}{3.04} & \ms{89.62}{3.58} & \ms{97.66}{0.72}
  & \ms{75.88}{3.21} & \ms{71.38}{4.17} & \ms{94.06}{1.01} \\
\midrule
TransMIL
  & \ms{92.97}{0.78} & \ms{92.41}{0.95} & \ms{96.16}{0.71}
  & \ms{86.59}{1.02} & \ms{85.21}{1.18} & \ms{91.62}{3.87}
  & \ms{84.73}{3.16} & \ms{80.17}{2.64} & \ms{94.18}{1.71}
  & \ms{69.70}{0.98} & \ms{63.77}{2.03} & \ms{91.85}{0.38} \\
MFE-TransMIL
  & \ms{94.79}{0.45} & \ms{94.44}{0.56} & \ms{98.39}{0.91}
  & \ms{91.29}{2.24} & \ms{90.37}{2.42} & \ms{93.17}{1.09}
  & \ms{94.06}{1.59} & \ms{90.95}{1.57} & \ms{97.67}{1.18}
  & \ms{70.61}{0.51} & \ms{63.93}{1.61} & \ms{92.57}{0.05} \\

\bottomrule
\end{tabular}
}
\end{table}

\section{Ablation Study}
\begin{figure}[!htbp]
  \centering
  \begin{subfigure}[t]{0.19\linewidth}
    \centering
    \includegraphics[width=\linewidth]{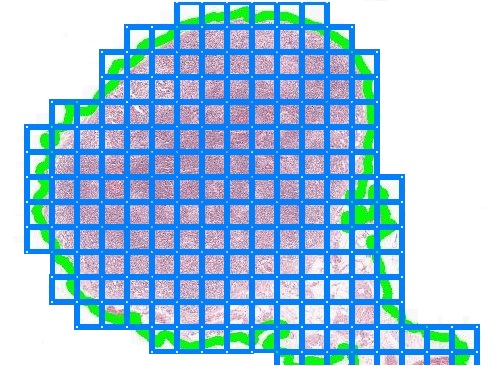}
    \caption{Original}
  \end{subfigure}\hfill
  \begin{subfigure}[t]{0.19\linewidth}
    \centering
    \includegraphics[width=\linewidth]{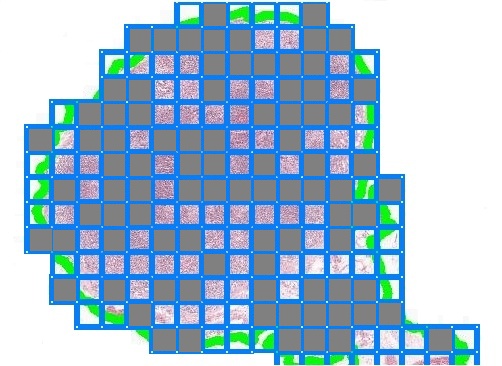}
    \caption{Single-token ($w\!=\!0$)}
  \end{subfigure}\hfill
  \begin{subfigure}[t]{0.19\linewidth}
    \centering
    \includegraphics[width=\linewidth]{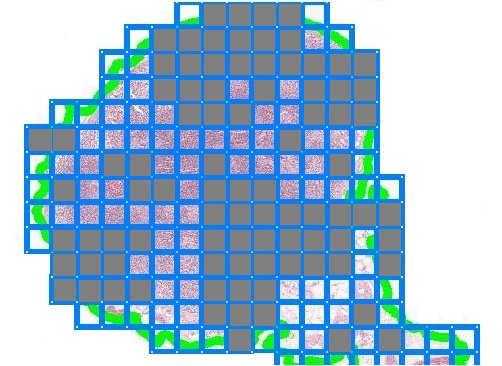}
    \caption{Small window ($w\!=\!1$, $3\!\times\!3$)}
  \end{subfigure}\hfill
  \begin{subfigure}[t]{0.19\linewidth}
    \centering
    \includegraphics[width=\linewidth]{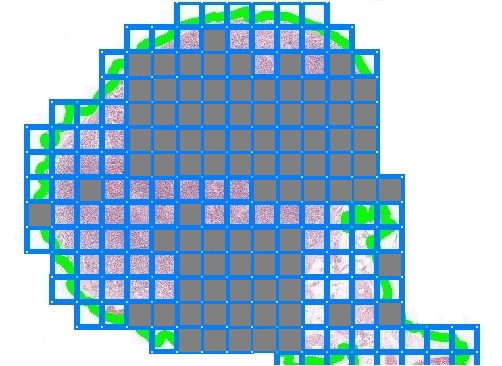}
    \caption{Large window ($w\!=\!2$, $5\!\times\!5$)}
  \end{subfigure}\hfill
  \begin{subfigure}[t]{0.19\linewidth}
    \centering
    \includegraphics[width=\linewidth]{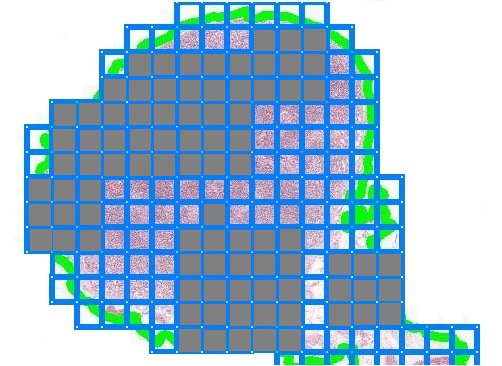}
    \caption{Adaptive ($r\!=\!75\%$)}
  \end{subfigure}
  \caption{Window-based adaptive masking ($r\!=\!75\%$), drawn on a spatial patch layout for
  readability; in training the mask is applied to the coordinate-free packed grid.
  Grey cells are masked tissue patches; green cells are background (absent tissue).
  Single-token masking ($w\!=\!0$) masks individual patches; small-window
  ($w\!=\!1$, $3\!\times\!3$) and large-window ($w\!=\!2$, $5\!\times\!5$)
  masking cover contiguous blocks of the grid to make reconstruction harder. The adaptive strategy places $3\!\times\!3$ and
  $5\!\times\!5$ windows (chosen at random) until they cover about 80\% of the
  masking budget, then fills the remainder with single cells.}
  \label{fig:masking}
\end{figure}

Table~\ref{tab:weight} reports the effect of the reconstruction-loss weight on PANDA.
\Cref{fig:masking} illustrates the window-based masking strategies compared in
the masking ablation of the main text: single-token, fixed small/large windows,
and the adaptive multi-scale schedule.

\begin{table}[!htbp]
\centering
\scriptsize
\setlength{\tabcolsep}{4pt}
\caption{Ablation on reconstruction-loss weight $\lambda_{rec}$. Results are AUC (\%) on PANDA across 3 splits $\times$ 5 seeds and five MFE-MIL variants based on MEAN, MAX, ABMIL (AB.), CLAM-MB (.MB), and CLAM-SB (.SB), plus average. \textbf{Bold} indicates the best average result.}
\setlength{\tabcolsep}{4pt}
\begin{tabular}{lcccccc}
\toprule
\textbf{$\lambda_{rec}$} & \textbf{MEAN+} & \textbf{MAX+} & \textbf{AB.+} & \textbf{.MB+} & \textbf{.SB+} & \textbf{Avg} \\
\midrule
\textbf{0.1} & 94.47 & 91.45 & 94.40 & 94.56 & 94.38 & 93.85 \\
\textbf{0.2} & 95.02 & 91.15 & 94.57 & 94.35 & 94.66 & 93.95 \\
\textbf{0.3} & 94.93 & 91.62 & 94.85 & 94.67 & 94.50 & \textbf{94.11} \\
\bottomrule
\end{tabular}
\label{tab:weight}
\end{table}
\section{Top-$k$ Attention Coherence (UNI)}
\label{sec:supp_overall_coherence}

This section provides the top-$k$ attention coherence summarised in the main
text for both datasets (Table~\ref{tab:topk_coherence}).

\begin{table}[!htbp]
\centering
\scriptsize
\caption{Top-$k$ attention coherence ($k\!=\!50$, ABMIL, UNI).
Higher gap = top-attended patches more cohesive than bottom-attended.}
\label{tab:topk_coherence}
\setlength{\tabcolsep}{4pt}
\begin{tabular}{@{}l ccc ccc@{}}
\toprule
 & \multicolumn{3}{c}{CAMELYON16} & \multicolumn{3}{c}{PANDA} \\
\cmidrule(lr){2-4}\cmidrule(lr){5-7}
Condition & top\_sim & cross\_sim & gap & top\_sim & cross\_sim & gap \\
\midrule
Frozen UNI & 0.370 & 0.187 & 0.183 & 0.496 & 0.309 & 0.187 \\
Baseline   & 0.908 & 0.678 & 0.230 & 0.785 & 0.563 & 0.222 \\
Rand MFE   & 0.922 & 0.670 & 0.252 & 0.765 & 0.544 & 0.221 \\
Win MFE    & 0.864 & 0.567 & \textbf{0.297} & 0.759 & 0.531 & \textbf{0.228} \\
\bottomrule
\end{tabular}
\end{table}

\section{PANDA Per-Grade Top-$k$ Attention Coherence}
\label{sec:supp_topk_panda}

Table~\ref{tab:topk_panda_grade} reports top-$k$ attention coherence
per Gleason grade on PANDA (1{,}031 test slides, $k=50$).
Grade~0 (benign) shows near-zero gap under all adapted conditions
($\leq\!0.039$) because benign slides lack a single focal lesion
and attention heads distribute mass broadly rather than concentrating
on a focal region.
Grades~1--5 are cancerous; Grades~2--5 show substantially larger gaps ($0.297$--$0.387$),
consistent with the adapter focusing attention on grade-defining tumour
glands.
Win MFE achieves the highest gap for Grades~0, 4, and 5, while Rand MFE
leads on Grade~1; on Grades~2 and~3 Baseline is slightly higher, so
differences between conditions are small and grade-dependent.

\begin{table}[!htbp]
\centering
\scriptsize
\caption{Per-grade top-$k$ attention coherence on PANDA ($k\!=\!50$,
1{,}031 test slides).
\textbf{top}: mean cosine similarity among the top-50 attended patches;
\textbf{cross}: similarity between top-50 and bottom-50 patches;
\textbf{gap} $=$ top $-$ cross.
Bold = highest gap per grade among the adapted conditions.}
\label{tab:topk_panda_grade}
\setlength{\tabcolsep}{3pt}
\resizebox{\linewidth}{!}{%
\begin{tabular}{@{}l r ccc ccc ccc ccc@{}}
\toprule
& & \multicolumn{3}{c}{\textbf{Frozen UNI}}
  & \multicolumn{3}{c}{\textbf{Baseline}}
  & \multicolumn{3}{c}{\textbf{Rand MFE}}
  & \multicolumn{3}{c}{\textbf{Win MFE}} \\
\cmidrule(lr){3-5}\cmidrule(lr){6-8}\cmidrule(lr){9-11}\cmidrule(lr){12-14}
Grade & $N$ & top & cross & gap
           & top & cross & gap
           & top & cross & gap
           & top & cross & gap \\
\midrule
Overall & 1031 & 0.496 & 0.309 & 0.187
               & 0.785 & 0.563 & 0.222
               & 0.765 & 0.544 & 0.221
               & 0.759 & 0.531 & \textbf{0.228} \\
Grade~0 &  278 & 0.508 & 0.311 & 0.197
               & 0.903 & 0.880 & 0.024
               & 0.882 & 0.854 & 0.028
               & 0.885 & 0.845 & \textbf{0.039} \\
Grade~1 &  262 & 0.528 & 0.313 & 0.215
               & 0.800 & 0.579 & 0.221
               & 0.795 & 0.549 & \textbf{0.246}
               & 0.759 & 0.550 & 0.209 \\
Grade~2 &  132 & 0.492 & 0.300 & 0.192
               & 0.745 & 0.403 & \textbf{0.342}
               & 0.705 & 0.398 & 0.307
               & 0.709 & 0.371 & 0.338 \\
Grade~3 &  121 & 0.473 & 0.314 & 0.159
               & 0.680 & 0.353 & \textbf{0.327}
               & 0.644 & 0.347 & 0.297
               & 0.649 & 0.323 & 0.326 \\
Grade~4 &  120 & 0.455 & 0.321 & 0.134
               & 0.700 & 0.389 & 0.311
               & 0.695 & 0.370 & 0.325
               & 0.692 & 0.359 & \textbf{0.333} \\
Grade~5 &  118 & 0.466 & 0.294 & 0.172
               & 0.708 & 0.350 & 0.358
               & 0.686 & 0.350 & 0.336
               & 0.702 & 0.316 & \textbf{0.387} \\
\bottomrule
\end{tabular}
}
\end{table}

\section{Effective Rank Analysis}
\label{sec:supp_effrank_panda}

Effective rank quantifies the intrinsic
dimensionality of a feature distribution as
$\text{eff\_rank} = \exp\!\bigl(-\sum_i p_i \log p_i\bigr)$,
where $p_i$ are the normalised singular values of the feature matrix.
Table~\ref{tab:eff_rank} reports effective rank on CAMELYON16
(128 test WSIs, overall and class-conditional) and
Table~\ref{tab:effrank_panda_grade} reports it per Gleason grade on
PANDA (1{,}031 test slides).

On both datasets the MIL classification objective is the dominant
compressor: frozen UNI features span a high-dimensional space
(eff\_rank $\approx\!53$ on CAMELYON16, $\approx\!39$--$41$ on PANDA),
and the Baseline adapter already collapses this to $21.8$ on CAMELYON16
and $\approx\!8$--$11$ on PANDA.
The reconstruction objective adds a further, dataset-dependent
compression: on CAMELYON16 it reduces eff\_rank appreciably beyond
Baseline ($21.8\!\to\!\approx\!17$), whereas on PANDA the three
conditions are within $\sim\!1$ of each other, so the additional effect
is negligible there.
Crucially, on \emph{both} datasets Rand MFE and Win MFE reach nearly
identical effective rank ($17.08$ vs $17.24$ on CAMELYON16; $10.36$ vs
$10.28$ on PANDA), so masking topology does not change dimensionality
reduction. This is consistent with the main-paper interpretation: the
masked-reconstruction objective, rather than the arrangement of the mask,
accounts for the extra compression on CAMELYON16, and block masking adds no
further dimensionality reduction over single-token masking.

\begin{table}[!htbp]
\centering
\scriptsize
\caption{Effective rank of adapter outputs on CAMELYON16 (UNI, 128 test WSIs).
Higher = more diverse directions preserved.}
\label{tab:eff_rank}
\setlength{\tabcolsep}{5pt}
\begin{tabular}{@{}lccc@{}}
\toprule
Condition & Overall & Normal & Tumor \\
\midrule
Frozen UNI & 52.86 & 53.04 & 52.57 \\
Baseline   & 21.80 & 22.32 & 20.96 \\
Rand MFE & 17.08 & 16.92 & 17.33 \\
Win MFE & 17.24 & 17.17 & 17.35 \\
\bottomrule
\end{tabular}
\end{table}

\begin{table}[!htbp]
\centering
\scriptsize 
\caption{Per-grade effective rank on PANDA (UNI, 1031 test slides).
Lower effective rank indicates more compressed feature representations.}
\label{tab:effrank_panda_grade}
\setlength{\tabcolsep}{6pt}
\begin{tabular}{@{}l r cccc@{}}
\toprule
Grade & $N$ & Frozen UNI & Baseline & Rand MFE & Win MFE \\
\midrule
Overall & 1031 & 39.44 &  9.78 & 10.36 & 10.28 \\
Grade~0 &  278 & 38.77 & 11.27 & 11.25 & 11.94 \\
Grade~1 &  262 & 38.71 &  8.97 &  9.30 &  9.48 \\
Grade~2 &  132 & 39.57 &  8.36 &  9.10 &  8.76 \\
Grade~3 &  121 & 39.85 &  9.49 & 10.84 &  9.90 \\
Grade~4 &  120 & 40.58 &  9.63 & 10.53 &  9.82 \\
Grade~5 &  118 & 40.91 & 10.16 & 11.38 & 10.67 \\
\bottomrule
\end{tabular}
\end{table}

\section{CONCH Encoder Spatial Coherence Analysis}
\label{sec:supp_conch}

Tables~\ref{tab:conch_coherence} and~\ref{tab:conch_entropy_topk} extend the
geometric analysis of the main paper to the CONCH encoder on 128 CAMELYON16
and 1{,}031 PANDA test WSIs.
Baseline results are omitted; only frozen CONCH versus Win MFE adapted features
are compared.

The overall compression pattern is consistent with UNI: the adj/rand ratio
drops on both datasets, confirming that adaptation induces semantic
compression regardless of encoder.
On PANDA, Grade~0 compresses most, as with UNI, while mid and high grades
change little: Grades~2--4 show a small ratio increase after adaptation.

The class-conditional direction on CAMELYON16 differs from UNI: tumour tissue
retains a slightly higher absolute gap than normal after adaptation, reversing the UNI ordering.
This is encoder-dependent behaviour and does not affect the primary finding
that compression preserves discriminability.

Attention entropy decreases on both datasets, consistent with UNI.
Top-$k$ coherence gap on PANDA increases, consistent
with UNI.
On CAMELYON16, the top-$k$ gap decreases because frozen
CONCH features already exhibit high coherence,
leaving less room for growth after compression; this is a ceiling effect rather
than a failure of the mechanism.

\begin{table}[!htbp]
\centering
\scriptsize 
\caption{Spatial coherence with CONCH encoder (frozen vs Win MFE).
\textit{Top}: overall and class-conditional on CAMELYON16 (128 WSIs).
\textit{Bottom}: overall and per-grade on PANDA (1{,}031 WSIs).
gap $=$ adj$-$rand.}
\label{tab:conch_coherence}
\setlength{\tabcolsep}{4pt}
\begin{tabular}{@{}llcccc@{}}
\toprule
Condition & Class & adj & rand & ratio & gap \\
\midrule
\multicolumn{6}{l}{\textit{CAMELYON16}} \\
Frozen CONCH & overall & 0.812 & 0.581 & $1.403\times$ & 0.231 \\
             & tumour  & 0.813 & 0.569 & $1.438\times$ & 0.245 \\
             & normal  & 0.811 & 0.589 & $1.381\times$ & 0.222 \\
\midrule
Win MFE      & overall & 0.961 & 0.897 & $1.075\times$ & 0.065 \\
             & tumour  & 0.953 & 0.872 & $1.099\times$ & 0.081 \\
             & normal  & 0.967 & 0.912 & $1.060\times$ & 0.054 \\
\midrule
\multicolumn{6}{l}{\textit{PANDA overall}} \\
Frozen CONCH & overall & 0.762 & 0.634 & $1.208\times$ & 0.128 \\
Win MFE      & overall & 0.808 & 0.699 & $1.173\times$ & 0.108 \\
\bottomrule
\end{tabular}

\smallskip

\begin{tabular}{@{}lrccc@{}}
\toprule
& & \multicolumn{3}{c}{adj/rand ratio} \\
\cmidrule(l){3-5}
Grade & $N$ & Frozen & Win MFE & $\Delta$ \\
\midrule
0 (benign)     & 278 & $1.190\times$ & $1.077\times$ & $-0.112$ \\
1              & 262 & $1.193\times$ & $1.146\times$ & $-0.048$ \\
2              & 132 & $1.219\times$ & $1.239\times$ & $+0.020$ \\
3              & 121 & $1.214\times$ & $1.256\times$ & $+0.041$ \\
4              & 120 & $1.224\times$ & $1.228\times$ & $+0.004$ \\
5 (aggressive) & 118 & $1.246\times$ & $1.242\times$ & $-0.004$ \\
\bottomrule
\end{tabular}
\end{table}

\begin{table}[!htbp]
\centering
\scriptsize 
\caption{Attention entropy and top-$k$ coherence with CONCH encoder
($k\!=\!50$, ABMIL). gap $=$ top\_sim $-$ cross\_sim.}
\label{tab:conch_entropy_topk}
\setlength{\tabcolsep}{5pt}
\begin{tabular}{@{}llcccc@{}}
\toprule
Dataset & Condition & Entropy & top\_sim & cross\_sim & gap \\
\midrule
CAMELYON16 & Frozen CONCH & 0.996 & 0.647 & 0.343 & 0.304 \\
           & Win MFE      & 0.842 & 0.832 & 0.598 & 0.234 \\
\midrule
PANDA      & Frozen CONCH & 0.953 & 0.678 & 0.456 & 0.222 \\
           & Win MFE      & 0.830 & 0.722 & 0.440 & 0.282 \\
\bottomrule
\end{tabular}
\end{table}

\section{Simple Normalization Baseline}
\label{sec:supp_norm}

We verify that the learned adapter is not interchangeable with a simple
per-slide feature transform.
Per-slide z-score normalization (centering and unit-variance scaling of patch
features before ABMIL) yields spatial ratio $= 2704.98$ and attention entropy
$= 0.982$ on CAMELYON16, indistinguishable from frozen UNI
($H\!\approx\!0.985$), because z-scoring collapses the common-mean structure
that cosine similarity relies on, driving random patch similarities near zero
and inflating the ratio spuriously.
On PANDA the ratio becomes negative ($-60.76$), confirming metric instability
under this transformation.
The learned MLP adapter is therefore not interchangeable with per-slide
whitening: it performs a task-aware projection that suppresses within-slide
variance while preserving discriminative geometry, an operation simple
normalisation cannot replicate.

\section{Packed-Grid Fidelity and Spatial-Mechanism Controls}
\label{sec:supp_controls}

\Cref{tab:fidelity} quantifies how well the coordinate-free packed pseudo-grid
recovers true spatial neighbours, and \cref{tab:controls} reports the
mask-arrangement controls referenced in the Geometric Analysis section of the main paper.

\begin{table}[!htbp]
\centering\scriptsize
\caption{Packed-grid fidelity vs.\ true patch coordinates (test slides).
prec@$k$: fraction of packed-grid neighbour pairs within $k$ true grid steps;
recall: fraction of true 8-neighbour pairs recovered; H/V-prec@1: within-row /
across-row precision@1. The pseudo-grid preserves only partial locality
(high within rows, near-zero across rows). Neighbours are measured directly from
exact patch coordinates one patch step (256) apart, without snapping to a grid.
CAMELYON16, PANDA and TCGA-BRCA use split 0 (128, 1{,}031 and 96 slides),
CAMELYON17 split 1 (50 slides); TCGA-BRCA uses 96 of 98 test slides (2 lacked
coordinate records).}
\label{tab:fidelity}
\setlength{\tabcolsep}{4pt}
\begin{tabular}{@{}lccccc@{}}
\toprule
Dataset & prec@1 & prec@2 & recall & H-prec@1 & V-prec@1 \\
\midrule
CAMELYON16 & 0.487 & 0.501 & 0.498 & 0.967 & 0.007 \\
CAMELYON17 & 0.482 & 0.495 & 0.499 & 0.957 & 0.007 \\
PANDA      & 0.469 & 0.515 & 0.500 & 0.894 & 0.043 \\
TCGA-BRCA  & 0.497 & 0.510 & 0.501 & 0.974 & 0.020 \\
\bottomrule
\end{tabular}
\end{table}

\begin{table}[!htbp]
\centering\scriptsize
\caption{Spatial-mechanism controls (UNI, ABMIL). All rows are a separate set of
matched re-runs ($\lambda_\text{rec}\!=\!0.3$, splits 0--2, seeds 1--5), so values
differ from the main result tables and the main-text coherence table. Coherence gap (adj$-$rand, mean over the 3 splits) and
test AUC (\%, mean over 3 splits $\times$ 5 seeds) on CAMELYON16/PANDA. The
arrangement mutations (shuffle, true-coordinate, 1D, distance-matched,
single-token) show no reliable ordering of the gap; standard window masking (Win)
has the highest CAMELYON16 AUC. Adapter (classification only) reference AUC is
$97.0$ (CAM16) / $94.2$ (PANDA).}
\label{tab:controls}
\setlength{\tabcolsep}{4pt}
\begin{tabular}{@{}lcccc@{}}
\toprule
Control & gap CAM16 & gap PANDA & AUC CAM16 & AUC PANDA \\
\midrule
Win (window MFE)         & 0.062 & 0.058 & 97.96 & 94.85 \\
Shuffle-order            & 0.059 & 0.054 & 95.40 & 94.27 \\
True-coordinate window   & 0.052 & 0.068 & 96.95 & 94.28 \\
1D contiguous            & 0.047 & 0.057 & 95.51 & 94.14 \\
Distance-matched random  & 0.061 & 0.061 & 97.38 & 94.21 \\
Single-token random      & 0.044 & 0.057 & 96.21 & 94.18 \\
\bottomrule
\end{tabular}
\end{table}

For reference, ACMIL (UNI) reaches AUC $97.1/93.6/97.4/93.9$ on
CAMELYON16/CAMELYON17/TCGA-BRCA/PANDA, below MFE-ABMIL ($99.3/95.1/98.2/94.9$)
on all four.

\section{t-SNE Visualisation}
\label{sec:supp_tsne}

\begin{figure}[!htbp]
    \centering
    \includegraphics[width=0.24\linewidth]{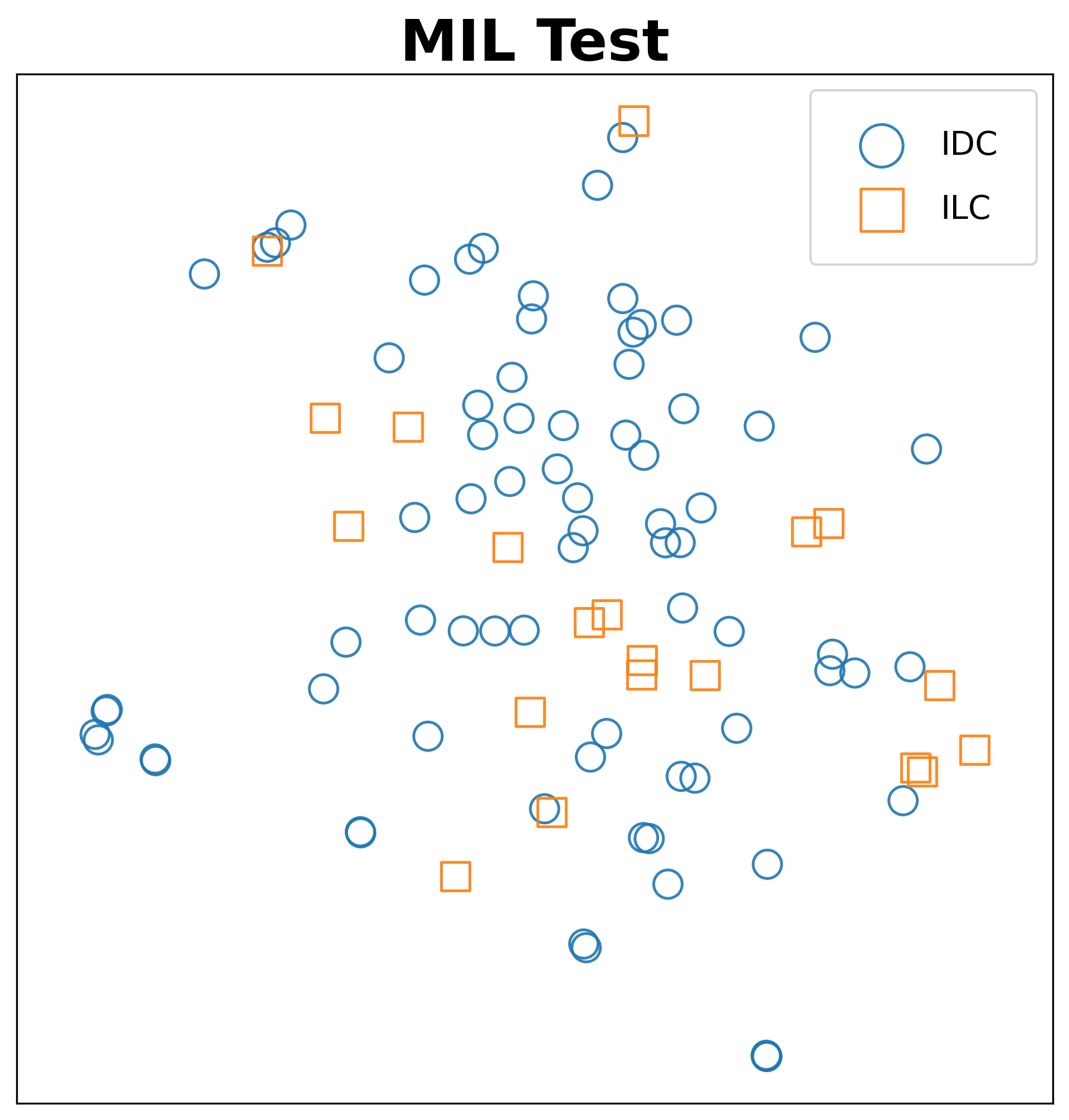}\hfill
    \includegraphics[width=0.24\linewidth]{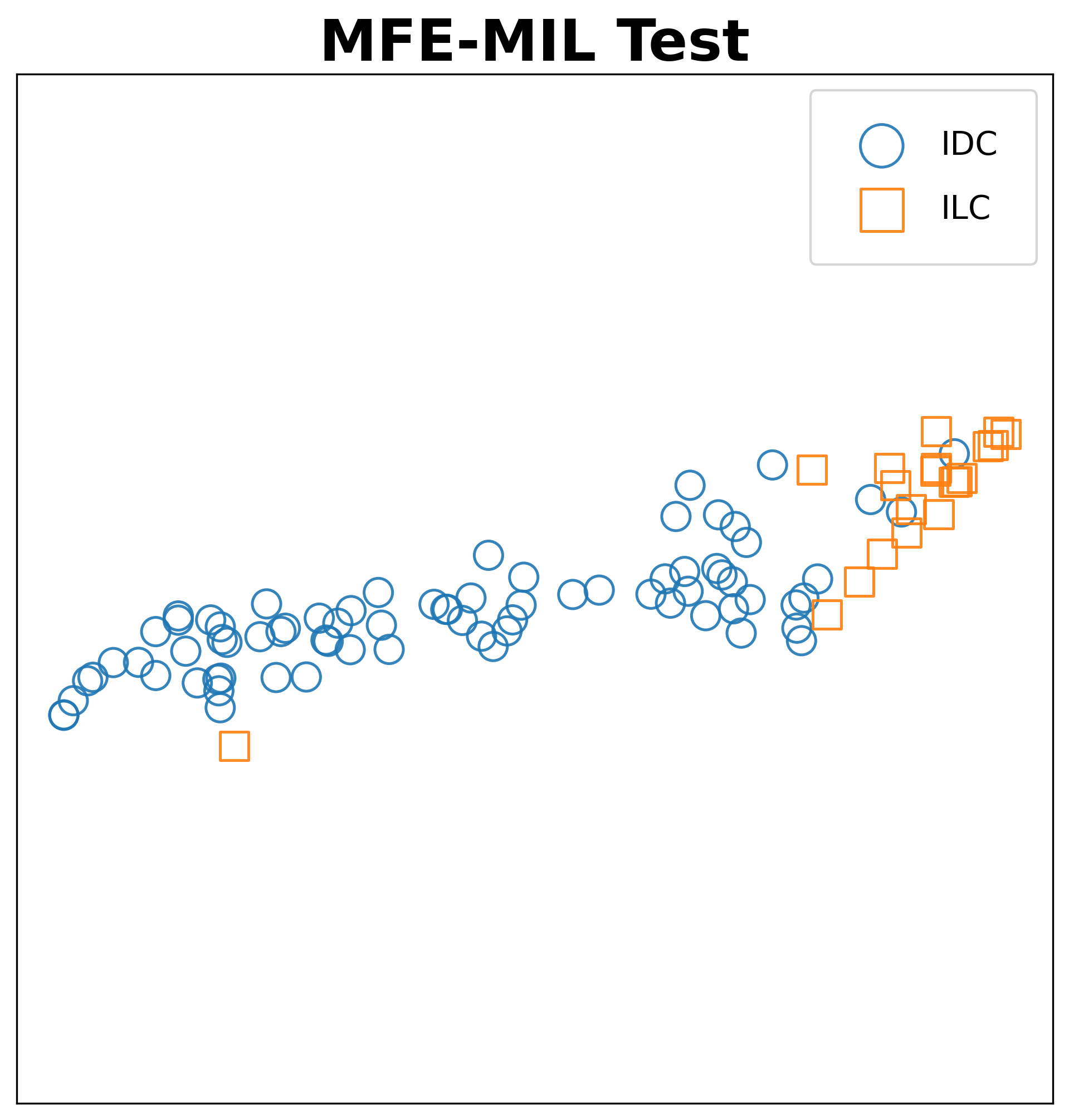}\hfill
    \includegraphics[width=0.24\linewidth]{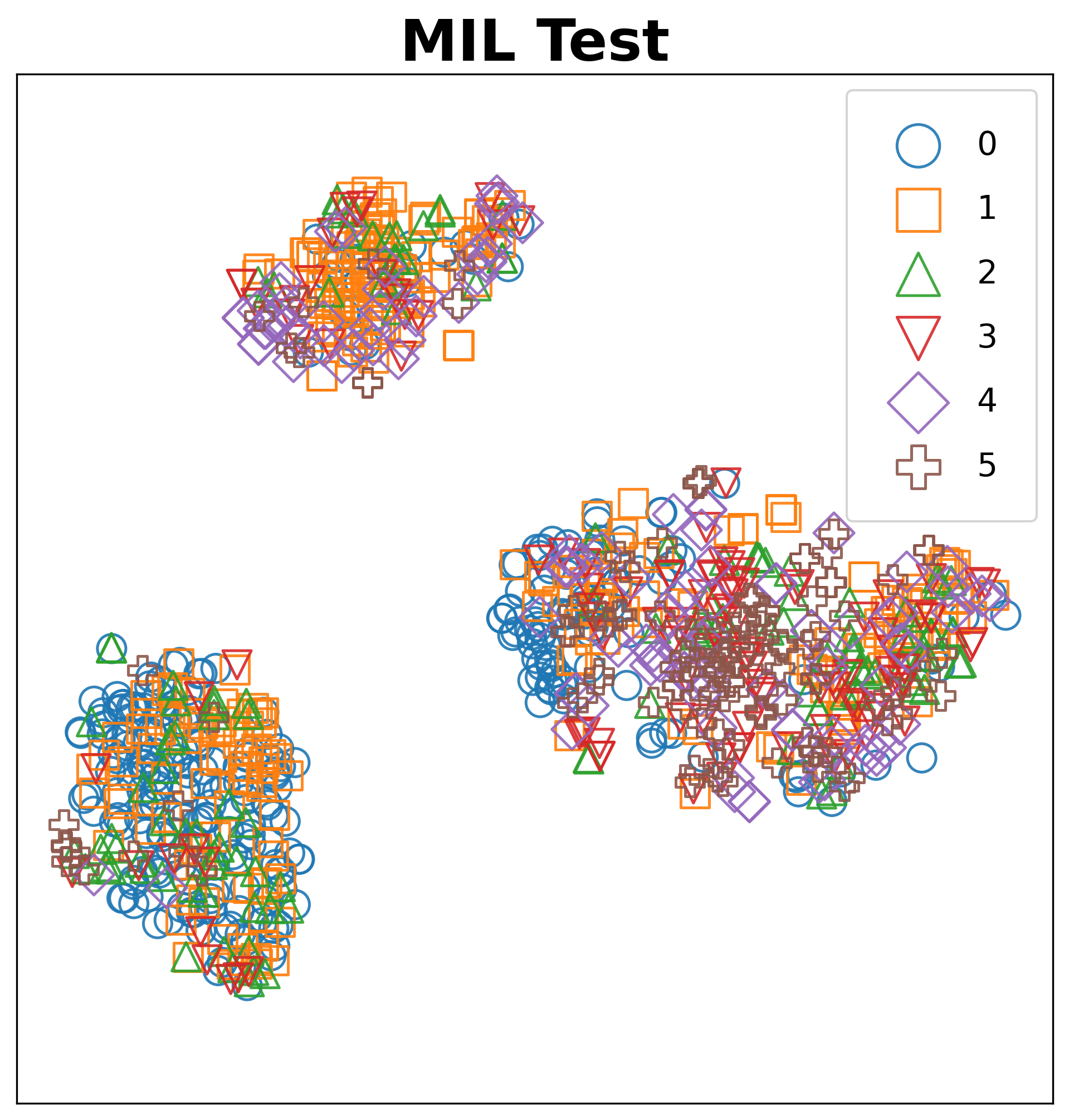}\hfill
    \includegraphics[width=0.24\linewidth]{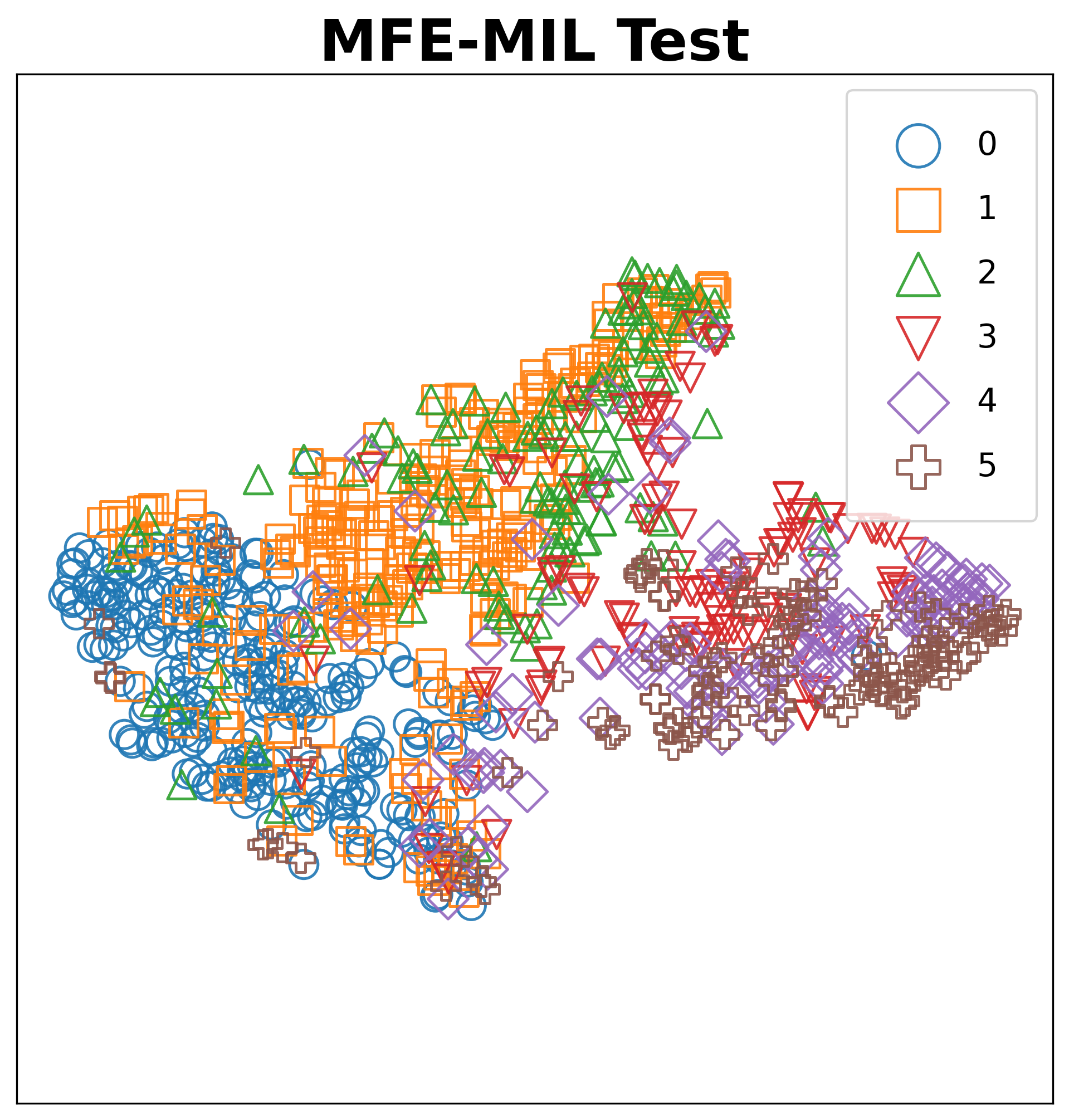}
    \caption{t-SNE of bag-level embeddings on the \textbf{test set}.
    From left to right: TCGA-BRCA MIL, TCGA-BRCA MFE-MIL, PANDA MIL, PANDA MFE-MIL.
    Each point represents one WSI bag, colored by class label. On TCGA-BRCA, MFE-MIL separates IDC and ILC into a clearly structured arc
    whereas MIL produces a fully mixed scatter.
    On PANDA, MFE-MIL reveals a visible ordinal gradient across Gleason grades (0--5)
    where MIL clusters all grades together without structure.}
    \label{fig:tsne}
\end{figure}

\begin{figure}[!htbp]
    \centering
    \includegraphics[width=0.24\linewidth]{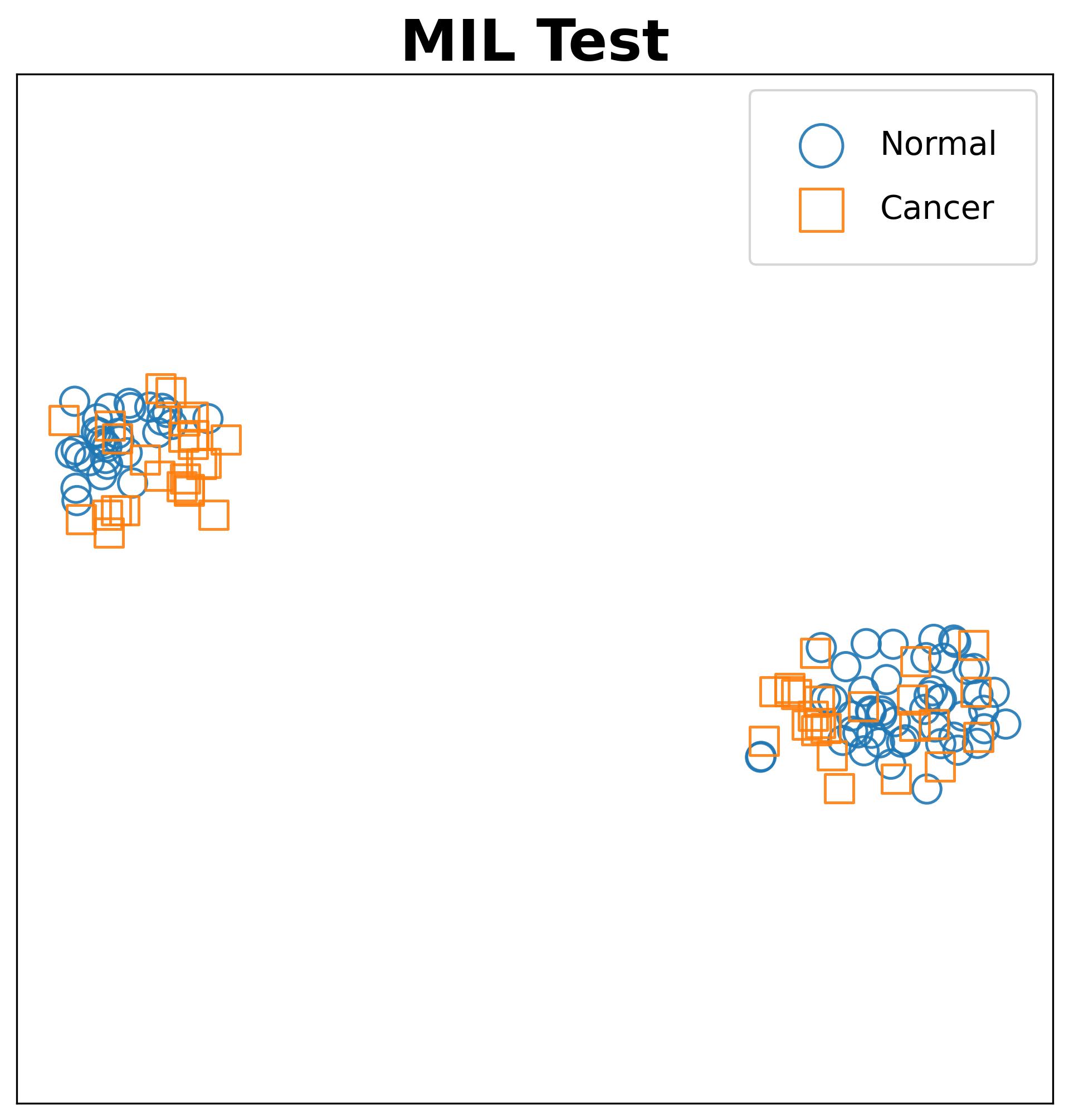}\hfill
    \includegraphics[width=0.24\linewidth]{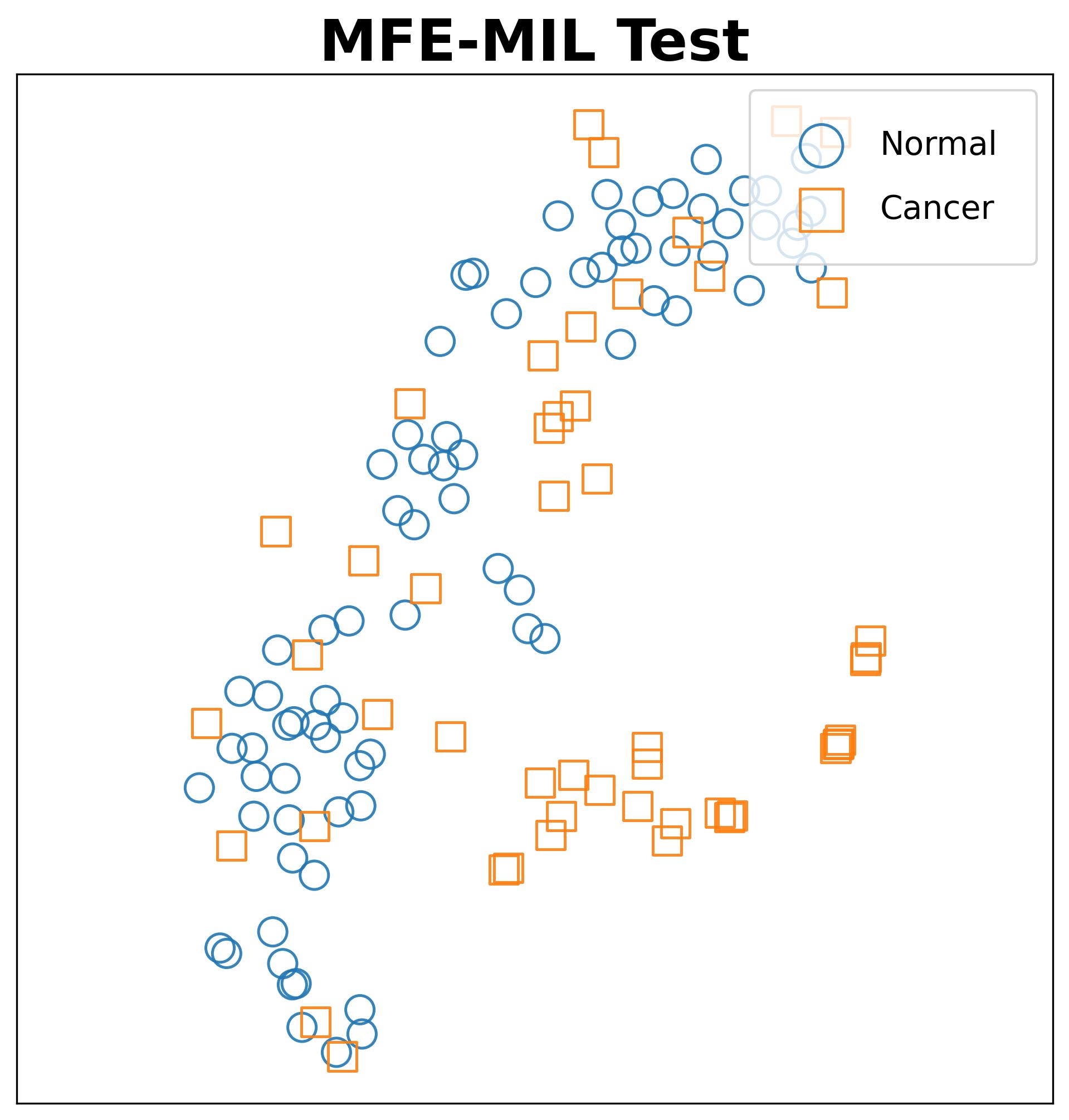}\hfill
    \includegraphics[width=0.24\linewidth]{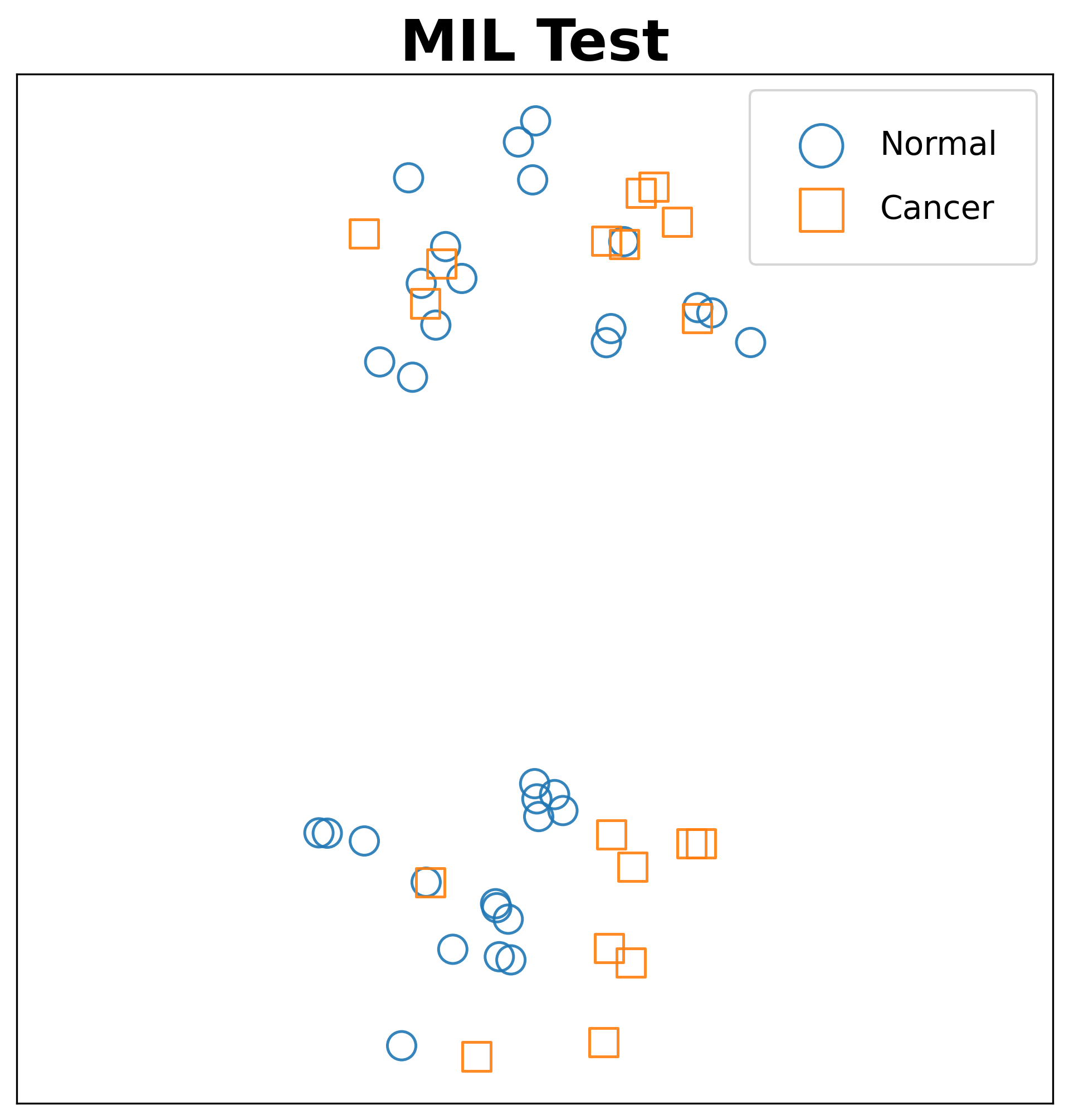}\hfill
    \includegraphics[width=0.24\linewidth]{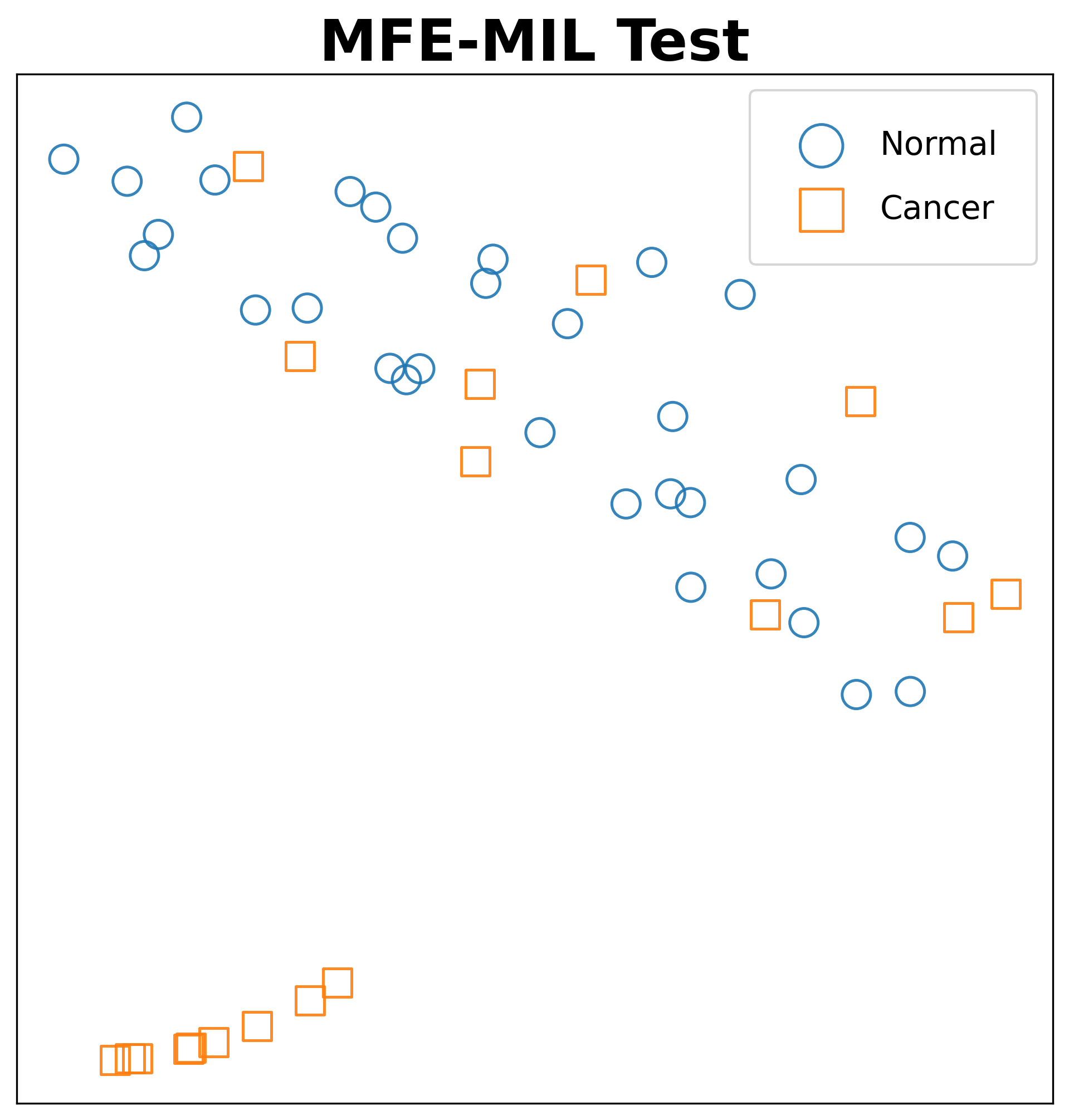}
    \caption{t-SNE of bag-level embeddings on the \textbf{test set}.
    From left to right: CAMELYON16 MIL, CAMELYON16 MFE-MIL, CAMELYON17 MIL, CAMELYON17 MFE-MIL.
    Each point represents one WSI bag, colored by class label. }
    \label{fig:tsne2}
\end{figure}

\section{Additional Qualitative Evaluation}
Attention heatmaps for one representative WSI per dataset, comparing each aggregator with and without MFE-MIL (UNI embeddings), followed by additional CAMELYON16 examples.

\begin{figure}[!htbp]
\centering
\newcommand{\colw}{0.135\linewidth}
\setlength{\tabcolsep}{2pt}
\begin{tabular}{>{\centering\arraybackslash}m{1.4cm} cccccc}
 &
\scriptsize\textbf{CAM16} &
\scriptsize\textbf{TCGA-KIRC} &
\scriptsize\textbf{TCGA-KIRP} &
\scriptsize\textbf{TCGA-LUAD} &
\scriptsize\textbf{TCGA-STAD} &
\scriptsize\textbf{TCGA-UCEC} \\[0pt]
\scriptsize\textbf{WSI} &
\includegraphics[width=\colw]{figs/heatmap/test_016_overlay.png} &
\includegraphics[width=\colw]{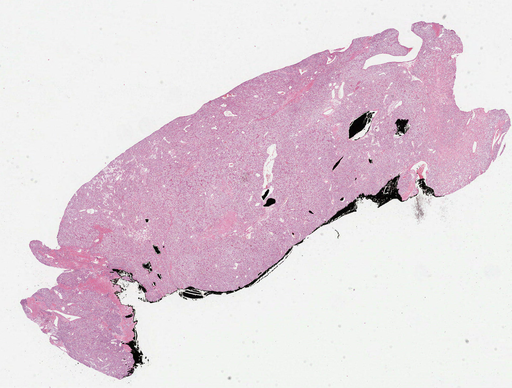} &
\includegraphics[width=\colw]{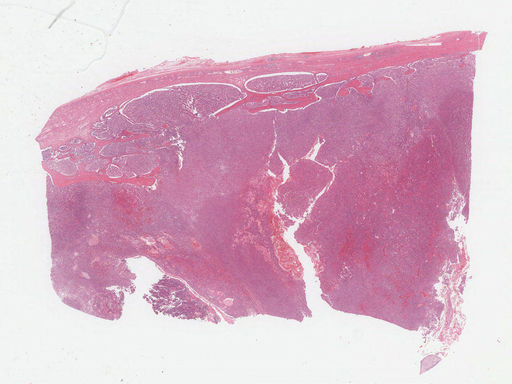} &
\includegraphics[width=\colw]{figs/LUAD/thumbnail/TCGA-05-4396-01Z-00-DX1.49DD5F68-7473-4945-B384-EA6D5AE383CB.png} &
\includegraphics[width=\colw]{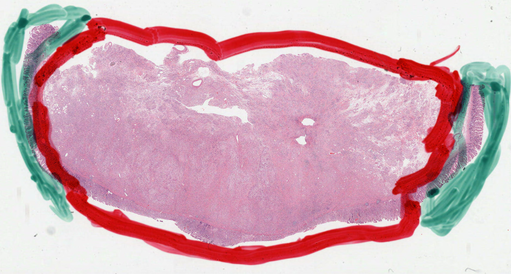} &
\includegraphics[width=\colw]{figs/heatmap/UCEC/thumbnail/TCGA-A5-A0GD-01Z-00-DX1.89F8C4CA-A364-4510-A8AE-34337AD761E9.png} \\[0pt]
\scriptsize\textbf{ABMIL} &
\includegraphics[width=\colw]{figs/heatmap/mil_att_test_016.png} &
\includegraphics[width=\colw]{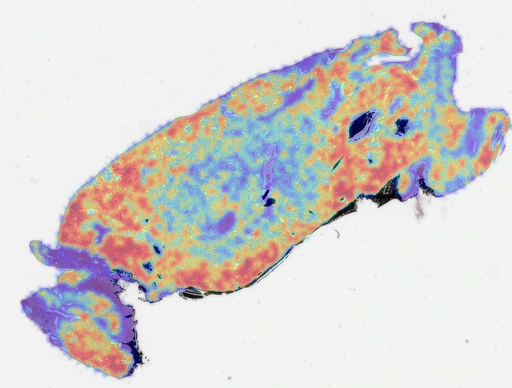} &
\includegraphics[width=\colw]{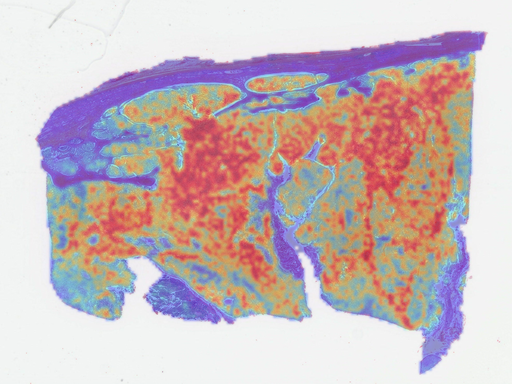} &
\includegraphics[width=\colw]{figs/LUAD/heatmaps/pure_mil/att/TCGA-05-4396-01Z-00-DX1.49DD5F68-7473-4945-B384-EA6D5AE383CB.png} &
\includegraphics[width=\colw]{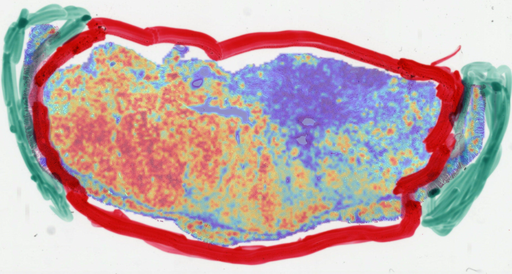} &
\includegraphics[width=\colw]{figs/heatmap/UCEC/heatmaps/pure_mil/att/TCGA-A5-A0GD-01Z-00-DX1.89F8C4CA-A364-4510-A8AE-34337AD761E9.png} \\[0pt]
\scriptsize\textbf{MFE-ABMIL} &
\includegraphics[width=\colw]{figs/heatmap/mfe_att_test_016.png} &
\includegraphics[width=\colw]{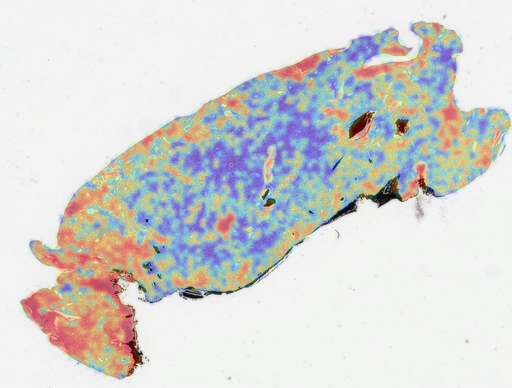} &
\includegraphics[width=\colw]{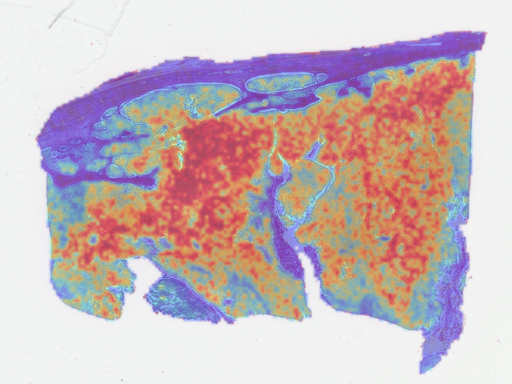} &
\includegraphics[width=\colw]{figs/LUAD/heatmaps/win_mae/att/TCGA-05-4396-01Z-00-DX1.49DD5F68-7473-4945-B384-EA6D5AE383CB.png} &
\includegraphics[width=\colw]{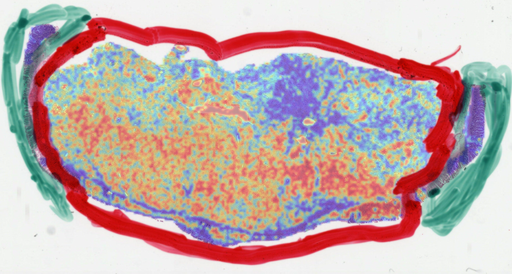} &
\includegraphics[width=\colw]{figs/heatmap/UCEC/heatmaps/win_mae/att/TCGA-A5-A0GD-01Z-00-DX1.89F8C4CA-A364-4510-A8AE-34337AD761E9.png} \\[0pt]
\scriptsize\textbf{CLAM-MB} &
\includegraphics[width=\colw]{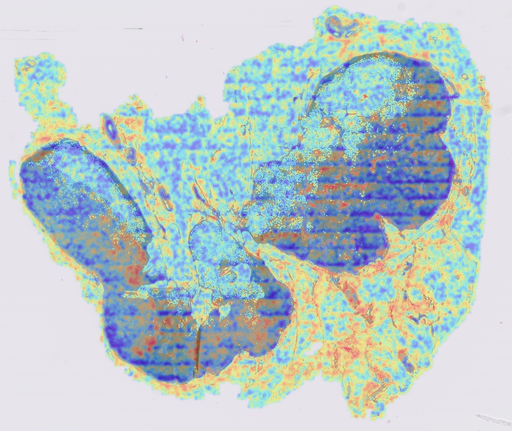} &
\includegraphics[width=\colw]{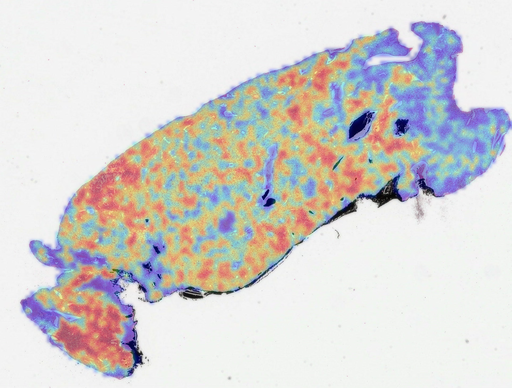} &
\includegraphics[width=\colw]{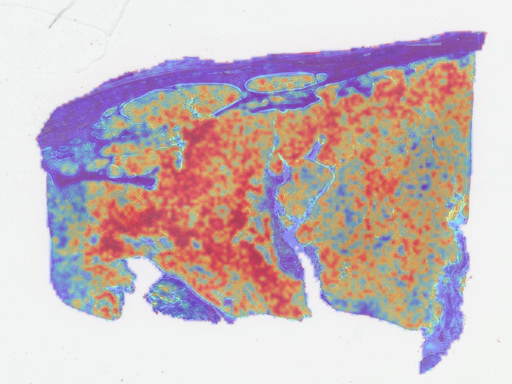} &
\includegraphics[width=\colw]{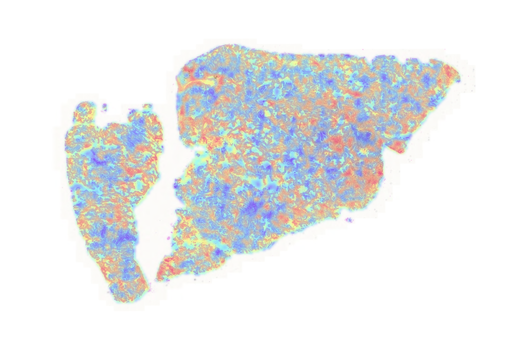} &
\includegraphics[width=\colw]{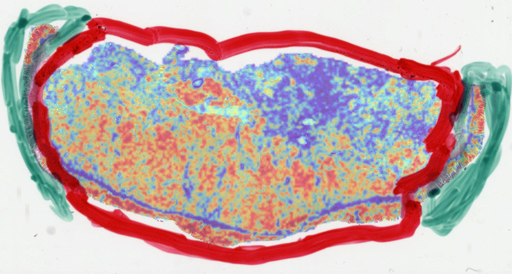} &
\includegraphics[width=\colw]{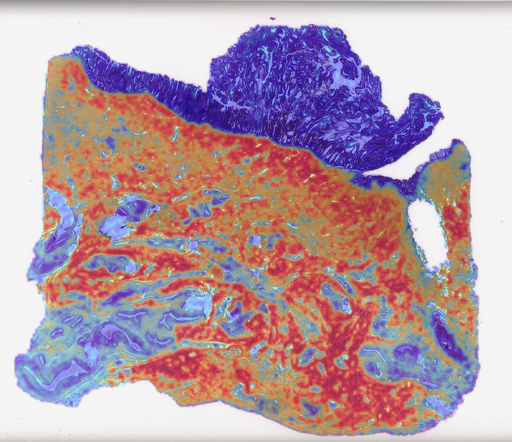} \\[0pt]
\scriptsize\textbf{MFE-CLAM-MB} &
\includegraphics[width=\colw]{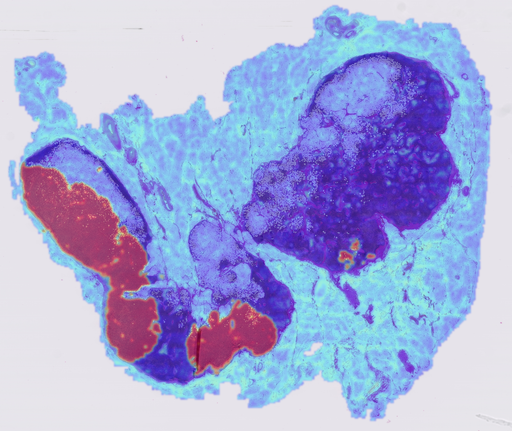} &
\includegraphics[width=\colw]{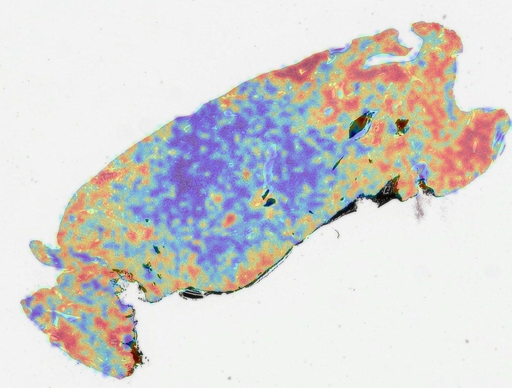} &
\includegraphics[width=\colw]{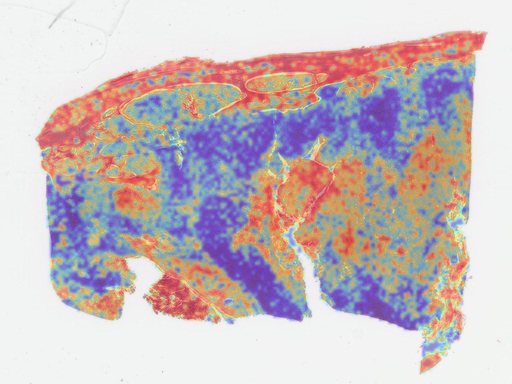} &
\includegraphics[width=\colw]{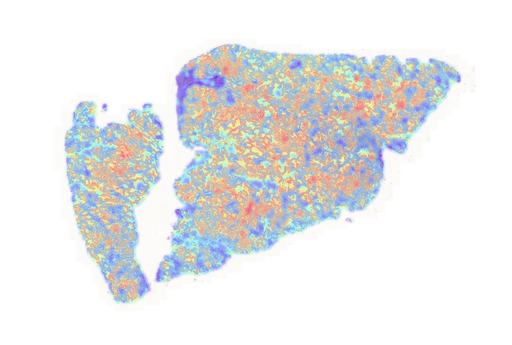} &
\includegraphics[width=\colw]{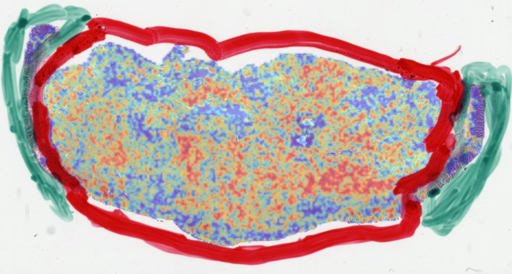} &
\includegraphics[width=\colw]{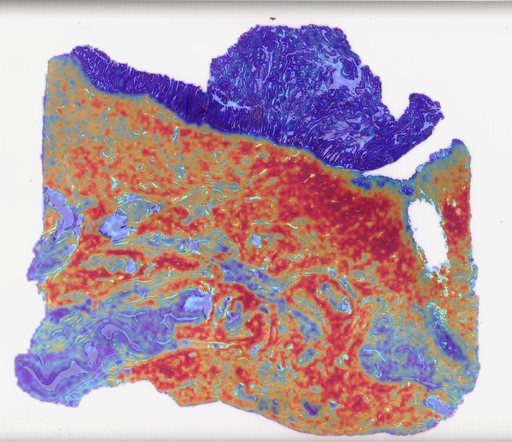} \\[0pt]
\scriptsize\textbf{CLAM-SB} &
\includegraphics[width=\colw]{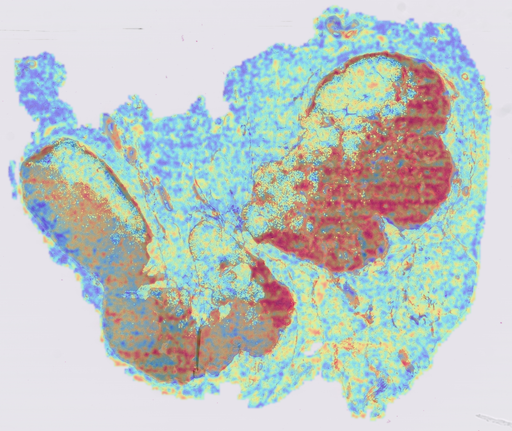} &
\includegraphics[width=\colw]{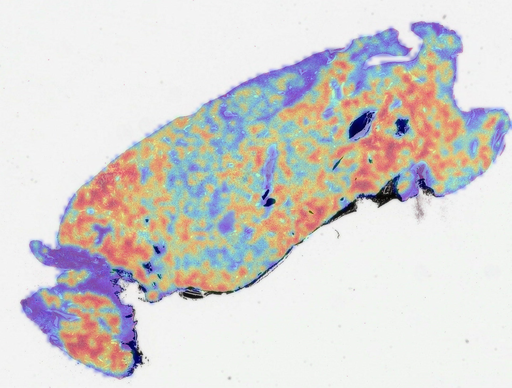} &
\includegraphics[width=\colw]{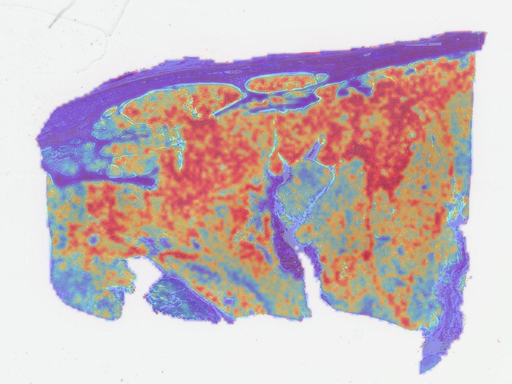} &
\includegraphics[width=\colw]{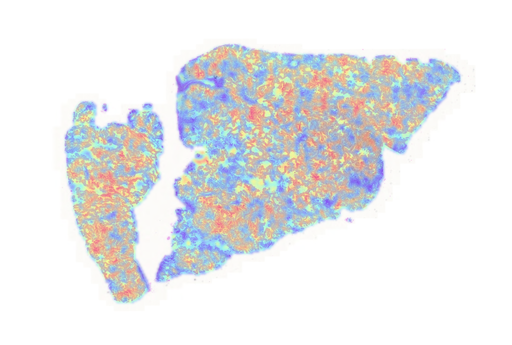} &
\includegraphics[width=\colw]{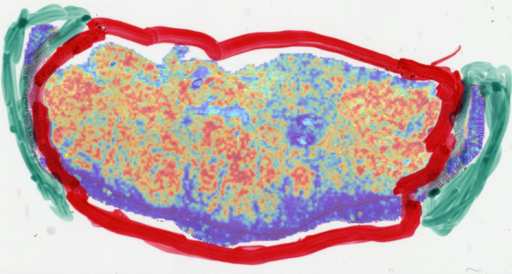} &
\includegraphics[width=\colw]{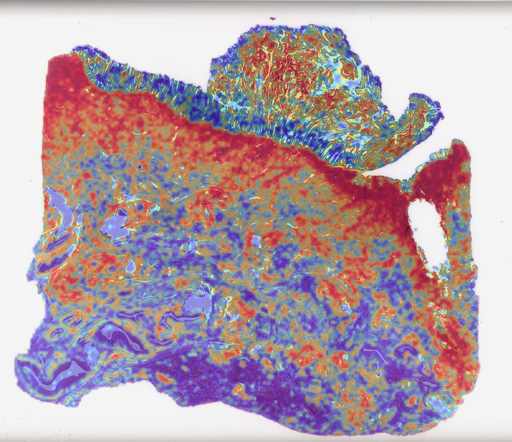} \\[0pt]
\scriptsize\textbf{MFE-CLAM-SB} &
\includegraphics[width=\colw]{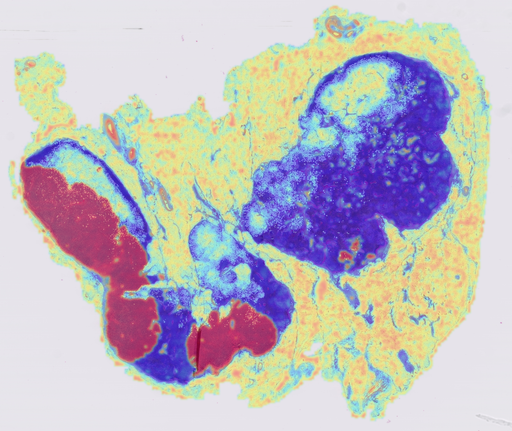} &
\includegraphics[width=\colw]{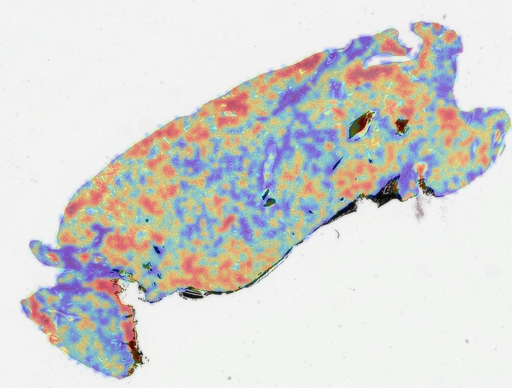} &
\includegraphics[width=\colw]{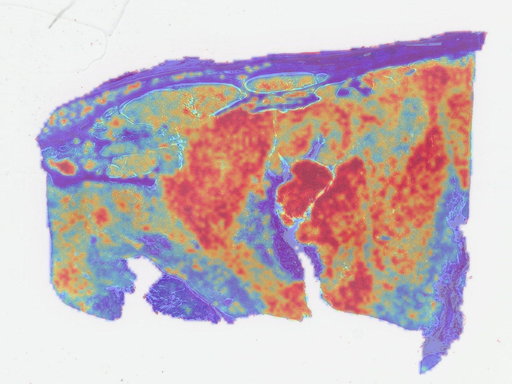} &
\includegraphics[width=\colw]{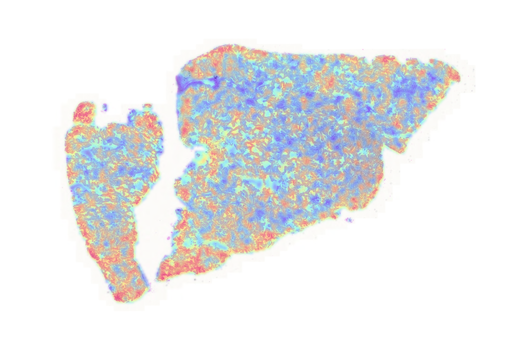} &
\includegraphics[width=\colw]{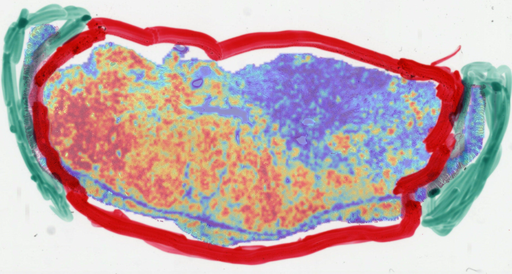} &
\includegraphics[width=\colw]{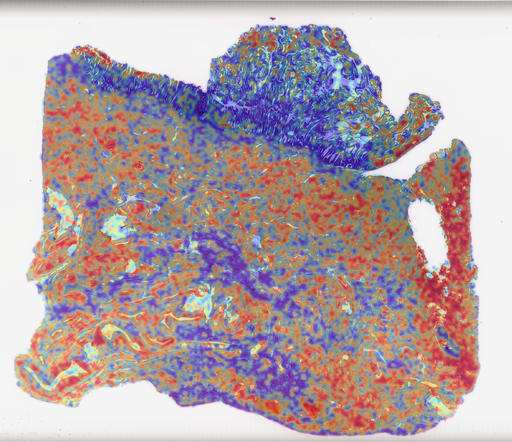} \\
\end{tabular}

\caption{Attention heatmaps: rows are WSI, ABMIL, MFE-ABMIL, CLAM-MB, MFE-CLAM-MB, CLAM-SB, MFE-CLAM-SB; columns are six datasets. Blue = low; red = high.
On CAMELYON16, baseline MIL concentrates attention on a tissue-fold artifact; MFE-MIL suppresses this and redirects to the annotated tumour.
In most cohorts, MFE-MIL redirects attention toward diagnostically relevant regions (KIRC and STAD are exceptions; see the main text).}
\label{fig:heatmaps}
\end{figure}

\begin{figure}[!htbp]
    \centering
        \begin{subfigure}[b]{0.28\textwidth} 
            \includegraphics[width=\textwidth]{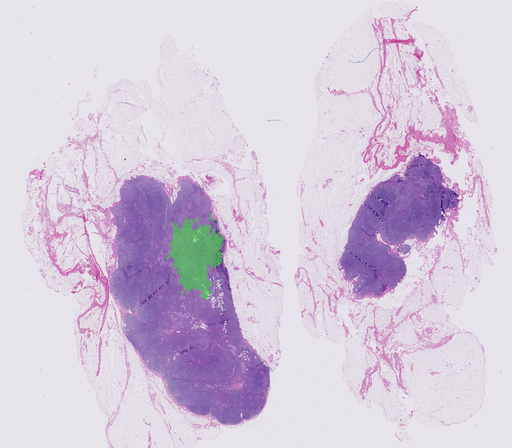}
        \end{subfigure}
        \begin{subfigure}[b]{0.28\textwidth}
            \includegraphics[width=\textwidth]{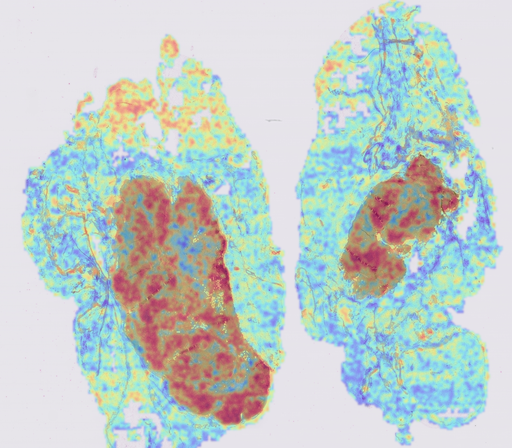}
        \end{subfigure}
        \begin{subfigure}[b]{0.28\textwidth}
            \includegraphics[width=\textwidth]{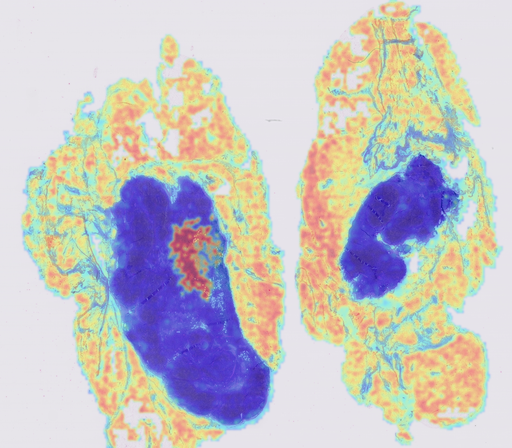}
        \end{subfigure}\\
        \begin{subfigure}[b]{0.28\textwidth} 
            \includegraphics[width=\textwidth]{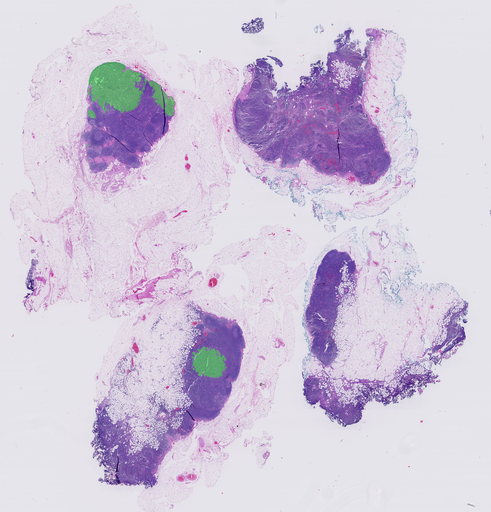}
        \end{subfigure}
        \begin{subfigure}[b]{0.28\textwidth}
            \includegraphics[width=\textwidth]{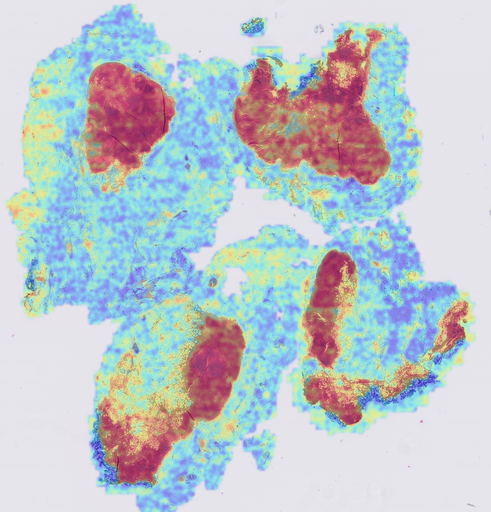}
        \end{subfigure}
        \begin{subfigure}[b]{0.28\textwidth}
            \includegraphics[width=\textwidth]{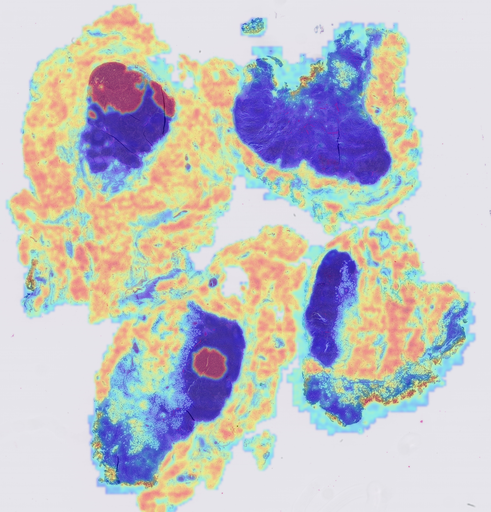}
        \end{subfigure}\\
        \begin{subfigure}[b]{0.28\textwidth} 
            \includegraphics[width=\textwidth]{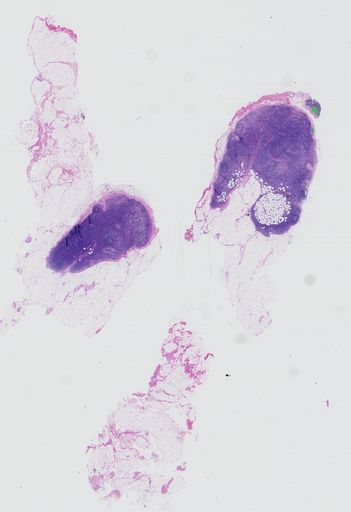}
        \end{subfigure}
        \begin{subfigure}[b]{0.28\textwidth}
            \includegraphics[width=\textwidth]{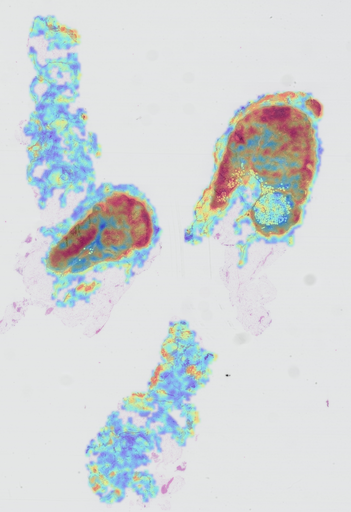}
        \end{subfigure}
        \begin{subfigure}[b]{0.28\textwidth}
            \includegraphics[width=\textwidth]{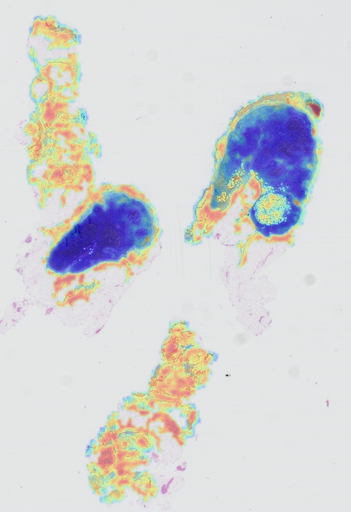}
        \end{subfigure}\\
        \begin{subfigure}[b]{0.28\textwidth} 
            \includegraphics[width=\textwidth]{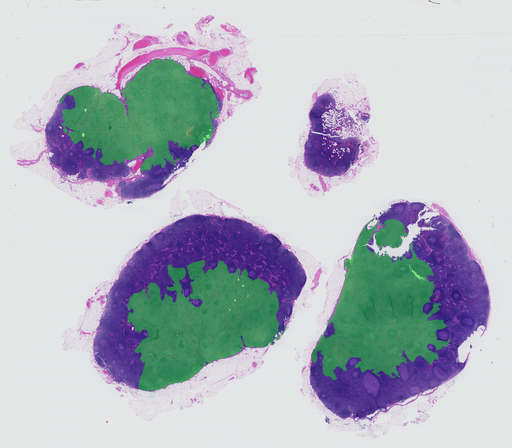}
            \caption{WSI}
        \end{subfigure}
        \begin{subfigure}[b]{0.28\textwidth}
            \includegraphics[width=\textwidth]{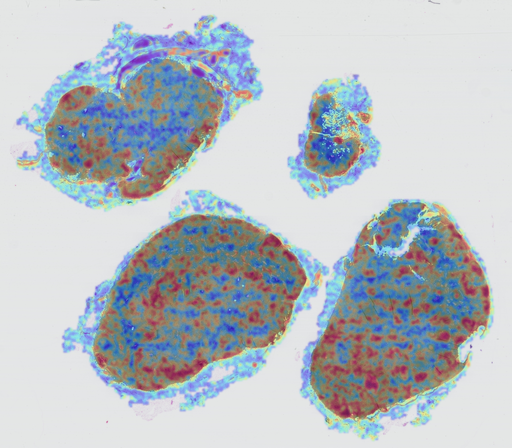}
            \caption{ABMIL}
        \end{subfigure}
        \begin{subfigure}[b]{0.28\textwidth}
            \includegraphics[width=\textwidth]{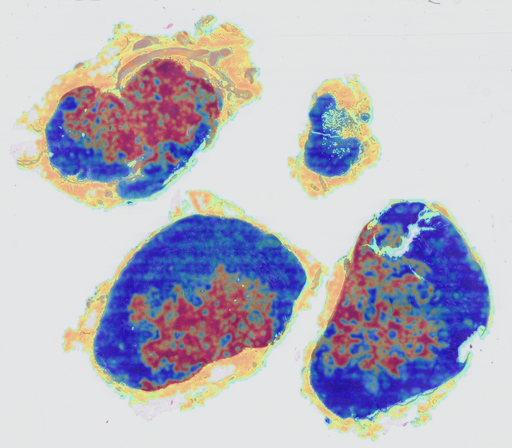}
            \caption{MFE-ABMIL}
        \end{subfigure}
        \caption{Attention maps of ABMIL and MFE-ABMIL on CAMELYON16 test 094, 001, 046 and 114 WSIs sample for cancer analysis. Tumor regions are annotated in green. In the right two columns, blue represents low attention (less important patches) and red represents high attention (critical patches). Compared to ABMIL, MFE-ABMIL more precisely focuses on regions critical for cancer detection.}
        \label{fig:heatmap2}
\end{figure}

\section{Limitations and Ethical Considerations}
\label{sec:supp_limits}

\noindent\textbf{Limitations.}
Packing patches in extraction order recovers spatial locality only approximately
(within raster rows, but with misalignment at row wraps and irregular tissue
boundaries; \cref{tab:fidelity}),
and the rare, small AUC drops (mostly on near-ceiling TCGA-BRCA) may reflect
compression removing variance that carries part of the signal, even as ACC/F1
improve.
AUC gains also shrink as the baseline strengthens, so MFE-MIL is most valuable for weaker-baseline aggregators, artifact suppression, and survival prediction.

\noindent\textbf{Ethical considerations.}
All experiments use publicly released, de-identified whole slide image datasets, namely CAMELYON16/17 (CC0), PANDA (CC BY-NC-SA 4.0), and TCGA (open-access data from the NIH Genomic Data Commons), which we used in accordance with their terms; no new patient data were collected. The qualitative attention-map assessment was a review of these public slides and model outputs by a board-certified pathologist and involved no patient contact or identifiable information. As only existing public, de-identified data were used, we did not seek separate ethics approval. MFE-MIL is a research method that has not been validated for clinical use, and any diagnostic deployment would require prospective validation and regulatory review.